\ifdefined\pdfoutput\pdfoutput=1\fi

\documentclass[sw]{iosart2x}

\usepackage{multirow}%
\usepackage{makecell}%
\usepackage{amsthm}
\usepackage{booktabs}%
\usepackage{algorithm}%
\usepackage{placeins}
\usepackage{algorithmicx}%
\usepackage{algpseudocode}%
\usepackage{anyfontsize}
\usepackage{rotating}
\usepackage[toc,page]{appendix} 

\usepackage{tabularx}
\usepackage{subcaption}
\usepackage{threeparttable}

\usepackage[framemethod=TikZ]{mdframed} 

\definecolor{iosblue}{RGB}{235, 245, 255} 

\newtheoremstyle{takeawaystyle}
  {0pt}{0pt}{\normalfont}{}{\bfseries}{.}{0.5em}{}

\theoremstyle{takeawaystyle}
\newtheorem*{takeawayinner}{Takeaway}

\newmdenv[
  roundcorner=5pt,
  linecolor=blue!20,
  linewidth=1pt,
  backgroundcolor=iosblue,
  innerleftmargin=10pt,
  innerrightmargin=10pt,
  innertopmargin=10pt,
  innerbottommargin=10pt,
  skipabove=1em,
  skipbelow=1em,
  nobreak=true
]{takeawayframe}

\newenvironment{takeaway_env}
  {\begin{takeawayframe}\begin{takeawayinner}}
  {\end{takeawayinner}\end{takeawayframe}}

\pubyear{2026}
\volume{0}
\firstpage{1}
\lastpage{1}
\makeatletter
\def\doc@stage{0}
\numberlinesfalse
\AtBeginDocument{%
  \def\copyright@price{}%
  \def\@copyrightowner{The authors.}%
  \def\standard@copyright@line{\copyright@statement}%
}
\makeatother

\begin{document}

\begin{frontmatter}

\title{Reranking Answers of Large Language Models with Knowledge Graphs}
\runtitle{Reranking Answers of Large Language Models with Knowledge Graphs}


\begin{aug}
\author[A,B]{\inits{M.}\fnms{Mikhail} \snm{Salnikov}\ead[label=e1]{Mikhail.Salnikov@skol.tech}}
\author[B]{\inits{H.}\fnms{Hai} \snm{Le}}
\author[C,D]{\inits{O.}\fnms{Olga} \snm{Tsymboi}}
\author[A,D]{\inits{I.}\fnms{Ivan} \snm{Lazichny}}
\author[B]{\inits{D.}\fnms{Dmitrii} \snm{Iarosh}}
\author[C,E]{\inits{E.}\fnms{Egor} \snm{Cheremiskin}}
\author[C,F]{\inits{A.}\fnms{Andrey} \snm{Savchenko}}
\author[C]{\inits{D.}\fnms{Dmitry} \snm{Simakov}}
\author[B,A]{\inits{A.}\fnms{Alexander} \snm{Panchenko}}
\address[A]{\orgname{Artificial Intelligence Research Institute}, Moscow, \cny{Russia}}
\address[B]{\orgname{Skolkovo Institute of Science and Technology}, Moscow, \cny{Russia}}
\address[C]{\orgname{Sber AI Lab}, Moscow, \cny{Russia}}
\address[D]{\orgname{Moscow Institute of Physics and Technology}, Moscow, \cny{Russia}}
\address[E]{\orgname{Moscow State University}, Moscow, \cny{Russia}}
\address[F]{\orgname{Laboratory for Theoretical Foundations of AI Models}, HSE University, Moscow, \cny{Russia}}
\end{aug}


\begin{abstract}
Answering natural language questions over knowledge graph data is challenging due to the vast number of facts, which can be difficult to process and navigate. One potential solution for this issue is to use mined subgraphs related to the query, although this process still requires extracting these subgraphs. This research presents a solution for extracting subgraphs related to entity candidates from a question-and-answer set, which can be obtained by inferring a large language model by calculating the shortest paths between entities. The proposed approaches detail various features that can be extracted from the subgraphs and reranking models to select the most probable answers from a list of candidates. Experiments were conducted on Wikidata to evaluate the effectiveness of the proposed approaches. This involved enumerating all the main feature types that can be extracted from mined subgraphs and a detailed analysis of the proposed features and reranking method combinations. In addition, a public web application that provides a useful web tool for studying the graph space between question and answer entities has been developed to work with subgraphs. This includes visualization of the extracted subgraph and automatic generation of natural language text to describe it.
\end{abstract}

\begin{keyword}
\kwd{Large Language Model}
\kwd{Knowledge Graph}
\kwd{Shortest Path}
\kwd{Question Answering}
\end{keyword}

\accepted{for the Semantic Web journal, tracking no.~4075-5289; not published}
\acceptedby{Guest Editors of the special issue on Large Language Models, Generative AI and Knowledge Graphs}

\end{frontmatter}




\section{Introduction}\label{sec1}

Answering factoid questions is a challenging task for any QA system, especially when the system does not have access to relevant knowledge from a Knowledge Graph~(KG) or other external source. Despite the lack of a guaranteed correct answer, people still use Large Language Models~(LLM) to address this challenge~\cite{DBLP:conf/coling/SenAS22-mintaka, DBLP:journals/aiopen/ZhangLFZ21}, especially since the release of InstructGPT~\cite{DBLP:conf/nips/Ouyang0JAWMZASR22}. Given a natural language question and a knowledge base, this paper aims to generate an answer given the context of the KG. In addition to addressing the LLMs themselves, leveraging external structured knowledge bases such as Wikidata~\cite{wikidata}, DBPedia~\cite{dbpedia}, and NELL~\cite{nell} can potentially boost the suboptimal performance of these language models.
These Knowledge Bases (Knowledge Graphs) are excellent examples of the implementation of Semantic Web principles. They demonstrate how ontologies, RDF~(Resource Description Framework), and SPARQL can be used to create complex, interconnected, and machine-readable knowledge systems. While LLMs are great at generating human-like text, Semantic Web technologies are better at providing structured, interpretable, and reusable knowledge. This makes them very useful for tasks that require precision and explainability. 


As an alternative, unstructured data, such as plain text, can enhance the performance of LLM-powered factoid question answering through in-context learning. However, these techniques, known as Retrieval-Augmented Generation~(RAG)~\cite{DBLP:conf/nips/LewisPPPKGKLYR020-rag}, may suffer from factual inaccuracies due to the use of untrusted general plain text sources. As a result, RAG platforms frequently require data source filtering and fine-tuning of the retrieval model for downstream applications to ensure accuracy or some other engineering hacks and tricks. Additionally, performance degradation may arise from either the controversially retrieved context or the model's inability to process context properly.
LLM-based systems have been observed to suffer from hallucinations in their answer generation and reasoning processes, which has limited their utility for Knowledge Graph Question Answering (KGQA) tasks~\cite{DBLP:journals/corr/abs-2403-01390}. Furthermore, there is a challenge with long-tail facts, as dealing with information related to less frequent or obscure entities is difficult for all KGQA systems~\cite{DBLP:journals/corr/abs-2405-06524}, and particularly for language models due to the limited knowledge available in training corpora about this type of facts. 
At the same time, state-of-the-art KGQA systems perform poorly on complex datasets~\cite{DBLP:conf/paclic/SalnikovLRNBMP23-originalpaper}.



\begin{figure}[htb]
  \centering
  \includegraphics[width=0.6\columnwidth]{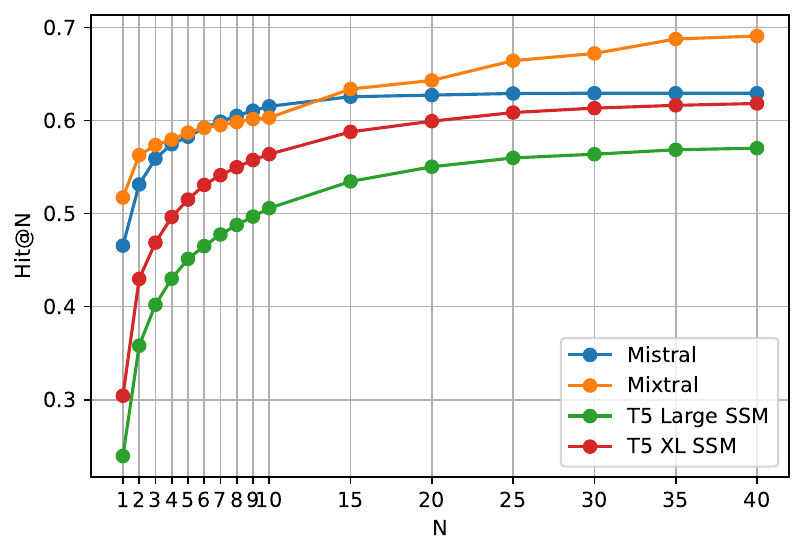}
  \caption{
      The motivation of our research: LLM's top-1 QA prediction accuracy is often incorrect. However, the correct answer is often in the top 10-40 alternative generations. Thus, a better reranking, such as the one presented in this article, is required.   Dependence of Hit@N metric on the size of answer pool $N$ on Mintaka~\cite{DBLP:conf/coling/SenAS22-mintaka} full dataset and Mistral~\cite{DBLP:journals/corr/abs-2310-06825-mistral}, Mixtral~\cite{DBLP:journals/corr/abs-2401-04088-mixtral}, T5-Large-SSM and T5-XL-SSM~\cite{DBLP:journals/jmlr/RaffelSRLNMZLL20-t5ssm} tuned models using Diverse Beam Search~\cite{vijayakumar2018diverse} as an inference strategy.  
  }
  \label{fig:hit_at_n_plot}
\end{figure}

Despite all problems, LLMs can answer factoid questions correctly in some cases, especially when prompted to generate multiple answers using some variation of Beam Search~\cite{vijayakumar2018diverse, DBLP:conf/konvens/SalnikovLBRMP23-actselection}. 
In other words, this shows that although the most likely option from the model's predictions can be incorrect, the answer pool can contain the correct answer. This highlights the importance of reranking answer pools and the use of accurate external data sources to improve the performance of a base language model after fine-tuning. 
As seen in Figure~\ref{fig:hit_at_n_plot}, where the {Hit@N} metric is evaluated for $1 \leq N\leq 40$, LLMs can generate accurate responses, at least for some beams. Therefore, if we have a method to select the correct responses, we can significantly improve the initial quality, bringing the performance closer to the ceiling. Our study will focus on the different reranking methods to enhance the base model's quality. 

This paper presents a collection of methods to address the KGQA challenge by accurately reranking possible answers using subgraphs. Firstly, we utilize LLM-produced responses that have previously demonstrated satisfactory outcomes on a complex factoid question-answering dataset, further described in Section~\ref{sec:base_model_expansion}. In addition, it has been shown that incorporation of the KG information into LLMs significantly improves the results for various Natural Language Processing task~\cite{DBLP:journals/tkde/PanLWCWW24}. Thus, this work focuses on extracting and utilizing the information about each question-answer pair by employing the information-dense Wikidata KG. To build upon the discussed novelty, we propose to look into this problem further in a reranking scope while still leveraging Wikidata as our external source. Our hypothesis for this study is that there is important information in the space between entities in the KG related to a question and its answer. According to the graph theory, such space can be represented as a subgraph. This information can help to justify the correctness of an answer. To investigate these subgraphs, we have developed a web application that allows us to explore them and study their relevance to the question-and-answer pair.

The contributions of our work are as follows:
\begin{enumerate}
    \item We propose a novel approach to the KGQA that utilizes subgraphs generated from paths from entities mentioned in a question to answer candidate entities. This approach is based on the observation that, while the correct answer may not always be the most likely according to language model predictions, it often appears later within the sequence of generated predictions. Our study consistently improves the Hit@N metric by using various features and models in KG-based reranking.

    \item Through our experiments, we have comprehensively compared the proposed method with well-known reranking techniques, ranging from classical algorithms based on graph features to more recent approaches using sentence transformers.

    \item A publicly available web application\footnote{\url{https://kgqa-nlp-zh.skoltech.ru/}} for KGQA, subgraph generation, visualization, and graph-to-text generation. This application demonstrates the relationships between the entities in the question and the answer, providing a better understanding of the graph's structure and motivation for one or another candidate to answer. 
\end{enumerate}

Our paper leverages extracted subgraphs and their features to rerank the language model (LM)'s generated answers. The main novelty compared to previous paper~\cite{DBLP:conf/paclic/SalnikovLRNBMP23-originalpaper} is defined as follows:
\begin{enumerate}
    \item The KGQA problem has been reformulated as a ranking problem, in contrast to previous work that focused on the top-1 answer, which does not provide sufficient information for a detailed understanding.
    
    \item We conducted in-depth experiments on all possible features and combinations that can be extracted from subgraphs to address the lack of such detailed experiments in previous studies.    

    \item The system incorporates graph-to-natural text generation features, which enhances the interpretability and usability of generated answers. 
    
    \item A subgraph visualization and graph-to-text representation have been added to the demo to understand better the graph's structure and the relationships between its entities.
    
\end{enumerate}

In this study, we have examined various aspects of working with subgraphs in KGQA and demonstrated that features derived from subgraphs can significantly enhance the ability of LLMs to answer questions. We release the publicly available source code on Github\footnote{\url{https://github.com/s-nlp/kbqa}} for reproducibility and transparent research.

\section{Background Works} \label{sec:related_work}

In this section, we provide an overview of the relevant research areas. 
Subsection~\ref{sec:kg} explores the KGs employed in this study. Next, in Subsection~\ref{sec:kgqa_methods}, we overview the key concepts of KGQA and discuss their advantages and limitations.
In Subsection~\ref{sec:factoid}, we discuss factoid questions and methods to determine answer correctness.

\subsection{Knowledge Graphs} \label{sec:kg}
KGs have emerged as powerful tools for organising and representing structured knowledge and facilitating intelligent applications. KGs consist of entities, relationships between these entities, and attributes that define the characteristics of each entity. These components form a semantic framework that allows data from multiple sources to be integrated and analysed. 
The concept of KGs originates from earlier theories, including semantic networks and ontologies. A typical KG is a collection of triples, each consisting of a subject, a relationship, and an object. The subject entity can be linked to other entities through specific relationships or properties defined by the relationship type.

\noindent There are several notable KGs:
\begin{enumerate}
    \item  Google KG: In 2012, Google introduced the KG, which significantly enhanced search capabilities by understanding the context and relationships between entities~\cite{singhal2012introducing}, marking a shift from keyword-based to entity-based search. This innovation improved the accuracy and relevance of search results; however, we are unfortunately unable to utilise this proprietary internal resource. The existence of KGs suggests that other industries may adopt similar structures, indicating potential for further research and development in this area.
    \item DBpedia: As a community-driven effort, DBpedia extracts structured information from Wikipedia and represents it as an RDF graph~\cite{DBLP:journals/ws/BizerLKABCH09}. It has become a central hub in the Linked Open Data (LOD) cloud, facilitating the integration and interoperability of diverse datasets.
    \item Wikidata: Launched by the Wikimedia Foundation, Wikidata is a collaborative knowledge base supporting Wikipedia and other projects within the Wikimedia community~\cite{wikidata}. It is a centralised repository for structured data, facilitating more sophisticated querying and analysis capabilities. Wikidata is a fascinating example of a general KG, encompassing millions of entities and a wide range of attributes.
\end{enumerate}

This paper focuses on the Wikidata KG, one of the most widely used general-purpose KGs. Their complexity makes the problem more challenging but also more realistic.

\subsection{Knowledge Graph Question Answering} \label{sec:kgqa_methods}

The KGQA task aims to generate accurate responses to a question based on facts extracted from KGs. 
KGQA systems utilise the structured data within KGs to provide accurate and context-aware responses to user inquiries. 
Leveraging language models has become crucial in various natural language processing applications, including KGQA. LLMs can perform zero-shot learning, producing answers for input prompts based on information stored in their pre-trained parameters, eliminating the need for additional training or labelled datasets.


Recent research has explored integrating knowledge from diverse sources—including unstructured documents, structured tables (e.g., from Wikipedia), and factual information from KGs—into language models~\cite{DBLP:conf/emnlp/KnowledgeAugmented}. The motivation for incorporating KGs into LMs lies in their ability to provide a compact yet rich representation of structured knowledge.

\subsection{Reranking in KGQA.}

Reranking candidates for QA is a widely recognized challenge that plays a crucial role in various applications, including information KG completion~\cite{DBLP:conf/coling/WangHHLYL24, DBLP:journals/corr/abs-2402-02389}. Initial candidate rankings often include noise or less relevant results, and reranking helps to refine these outputs, enhancing both their accuracy and relevance. The need for reranking in various applications further motivated us to apply this approach to KGQA with subgraphs, which is one of the novelties of this work.

Early studies in KGQA emphasized the importance of ranking for selecting structured evidence. \cite{Zhao_2019} showed that joint subject–relation scoring enhances subgraph selection on SimpleQuestions, while \cite{sun-etal-2024-efficient} demonstrated that entity-grouped retrieval combined with contextual triple-level reranking improves prior methods. These results highlight the role of KG ranking.

Recent approaches have integrated ranking directly into KG–LLM pipelines. For example, Q-KGR \cite{zhang-etal-2024-question} utilizes question-guided re-scoring to filter out irrelevant subgraph edges before incorporating LLMs. KG-Rank \cite{yang-etal-2024-kg} employs multi-stage triples reranking for medical QA, and KGR \cite{Guan_Liu_Lin_Lu_He_Han_Sun_2024} retrofits LLM answers by validating extracted statements against KGs, effectively reducing hallucinations. A recent survey \cite{ma-etal-2025-large-language-models-meet} categorizes these methodologies, highlighting the essential role of ranking in filtering KG information before or during interaction with the LLM.

Furthermore, retrieve-rerank architectures demonstrate their value beyond QA, particularly in knowledge graph completion. KGR$^3$ \cite{li-etal-2025-retrieval} merges embedding-based candidates with those generated by LLMs through dedicated reranking, resulting in improved missing entity prediction. ReranKGC \cite{GAO2025107467} employs CLIP-based reranking across multi-modal attributes, consistently boosting performance in the link prediction task.

\subsection{LLM-based reranking}
A parallel line of work applies LLMs directly as rankers to information retrieval. RankGPT \cite{sun-etal-2023-chatgpt} shows that zero-shot LLM re-rankers can outperform supervised rankers on MS MARCO and BEIR benchmarks. In \cite{10.1145/3626772.3657813} authors propose setwise prompting that jointly considers candidate sets, improving efficiency and effectiveness over pointwise, pairwise, and listwise approaches. REARANK extends this by formulating reranking as agentic listwise reasoning with reinforcement learning, achieving GPT-4-level performance with a 7B model \cite{zhang-etal-2025-rearank}. These works demonstrate LLMs can function as powerful inference-time rankers over unstructured candidates.

\subsection{Self-Consistency and Majority Vote}
A simple yet effective approach for answer selection is majority voting, which has been widely popularized in the large language model (LLM) domain as "self-consistency"~\cite{wang2022self}. This method leverages the idea that while a model may generate diverse incorrect answers, correct reasoning paths often converge on a single unique result.

By sampling multiple output candidates (e.g., through temperature sampling or beam search) and combining them through majority voting, the system can eliminate stochastic errors and select the most consistent response. Originally proposed for chain-of-thought reasoning, this approach can also serve as a robust baseline for factoid question answering (QA) by treating the frequency of a candidate answer in the generated list as a proxy for its likelihood.

\subsection{Factoid Questions} \label{sec:factoid}

Users commonly ask factoid questions, such as: (i) \textit{Which author wrote the most best-selling books in the decade that the internet was invented?} (ii) \textit{Who is the grandmother of the eldest grandchild of the current British monarch?}. 
Factoid questions usually have precise answers from a knowledge source such as KG or text corpus. We work with factoid questions and KGs as a source of knowledge in KGQA, a system for answering such questions. Answering factoid questions can be challenging without access to a KG, but KGQA systems that use KGs can achieve higher accuracy than traditional language models~\cite{DBLP:conf/emnlp/GuptaC018}. While language models can generate answers to factoid questions, they often produce incorrect responses, as illustrated in Figure~\ref{fig:hit_at_n_plot}. We hypothesise this is due to the lack of structured knowledge in KGs. Indeed, language models are trained on text, which may lack the same level of detail and accuracy as a regularly updated KG that can be edited.

\begin{figure}[htb]
  \centering
  \includegraphics[width=\columnwidth]{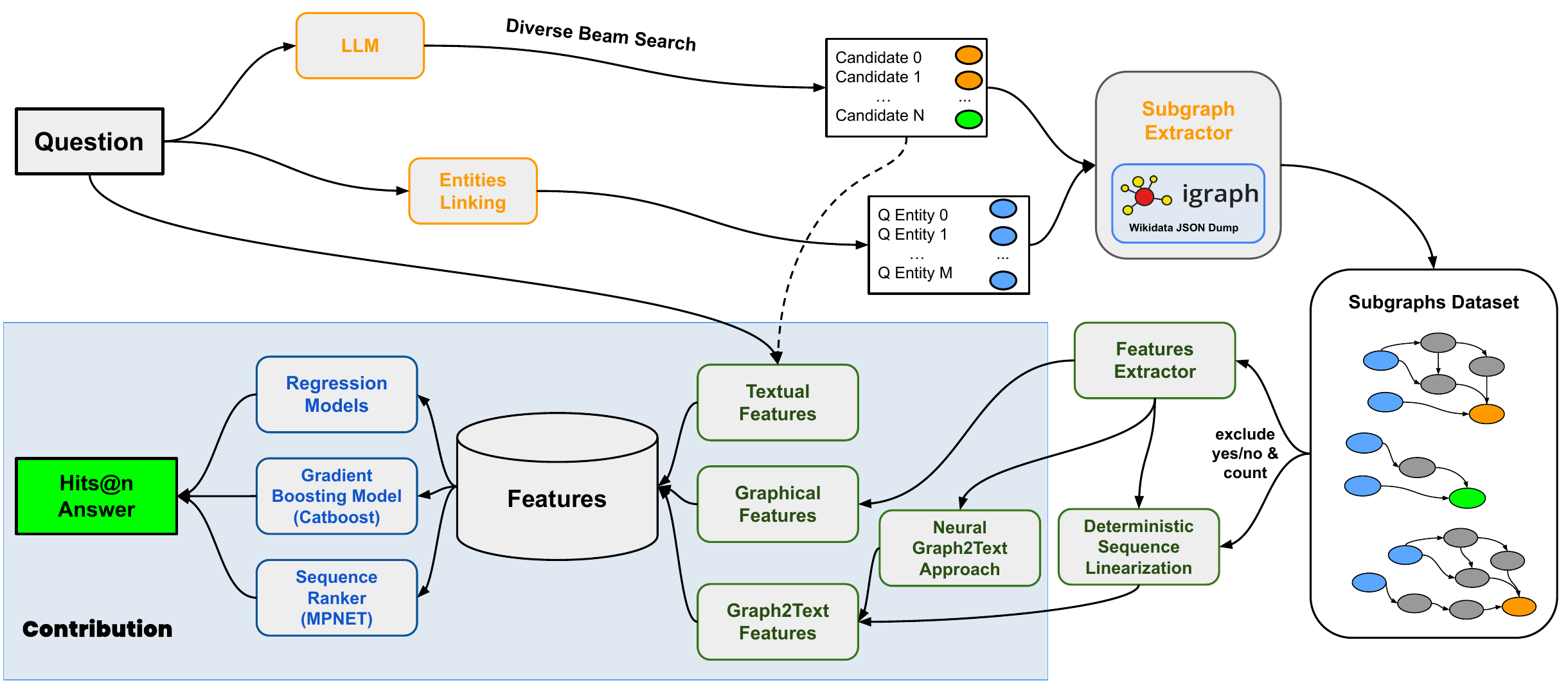}
  \caption{The proposed method for reranking language model answers with KGs. The method includes subgraph extraction, features extraction, and various ranker approaches. The subgraphs consist of the shortest paths between question entities and answer candidates, as discussed in Section~\ref{sec:subgraph_extract}. Various features were extracted for the ranker approaches, including text, graph, and Graph2Text Sequence Features, as discussed in Section~\ref{sec:subgraph_features}.}
  \label{big_pipe}
\end{figure}

\section{Methods}

In this study, we aim to explore the effectiveness of Knowledge Graph subgraphs in reranking LLM-generated answer candidates for factoid question answering. We formulate this task as follows: given a natural language question $Q$, a set of candidate answers $C=\{a_1, a_2, .., a_n\}$ produced by an LLM via Diverse Beam Search~\cite{vijayakumar2018diverse}, and a Knowledge Graph $G$. Our goal is to learn a scoring function $f(Q,a_i,G_i) \to \mathbb{R}$ that assigns higher scores to correct answers based on information from a dynamically constructed subgraph $G_i\subset G$.
Each subgraph $G_i$ is constructed by the union of all the shortest paths connecting the entities mentioned in the question with the candidate entity $a_i$. We hypothesize that subgraphs with correct answers provide important information for decision-making and should be crucial features for ranking function $f$.

To test this hypothesis, we propose three stages: (1) Subgraph Extraction, where we construct $G_i$ for each candidate; (2) Feature Engineering, where we prepare both graph features (e.g., density, path length) and textual representations for subgraph via Graph2Text linearization methods; and (3) Reranking, where these features are fed into learning-to-rank models to select the most accurate answer. The overview of the proposed pipeline is depicted in Figure~\ref{big_pipe}. The following subsections detail each stage.

\begin{takeaway_env}
Graph-Guided Reranking Approach that validates LLM generated answers by grounding them in Knowledge Graphs.
For every candidate answer generated by the LLM, we extract a specific subgraph connecting it to the question entities. This subgraph is then converted into a multi-modal feature vector, combining graph features with linearized "graph-to-text" narratives, which is fed into a learning-to-rank model to identify the most factually accurate answer.
\end{takeaway_env}

\subsection{Subgraph Extraction} \label{sec:subgraph_extract}
\begin{figure}[htb]
  \centering
  \includegraphics[width=0.7\columnwidth]{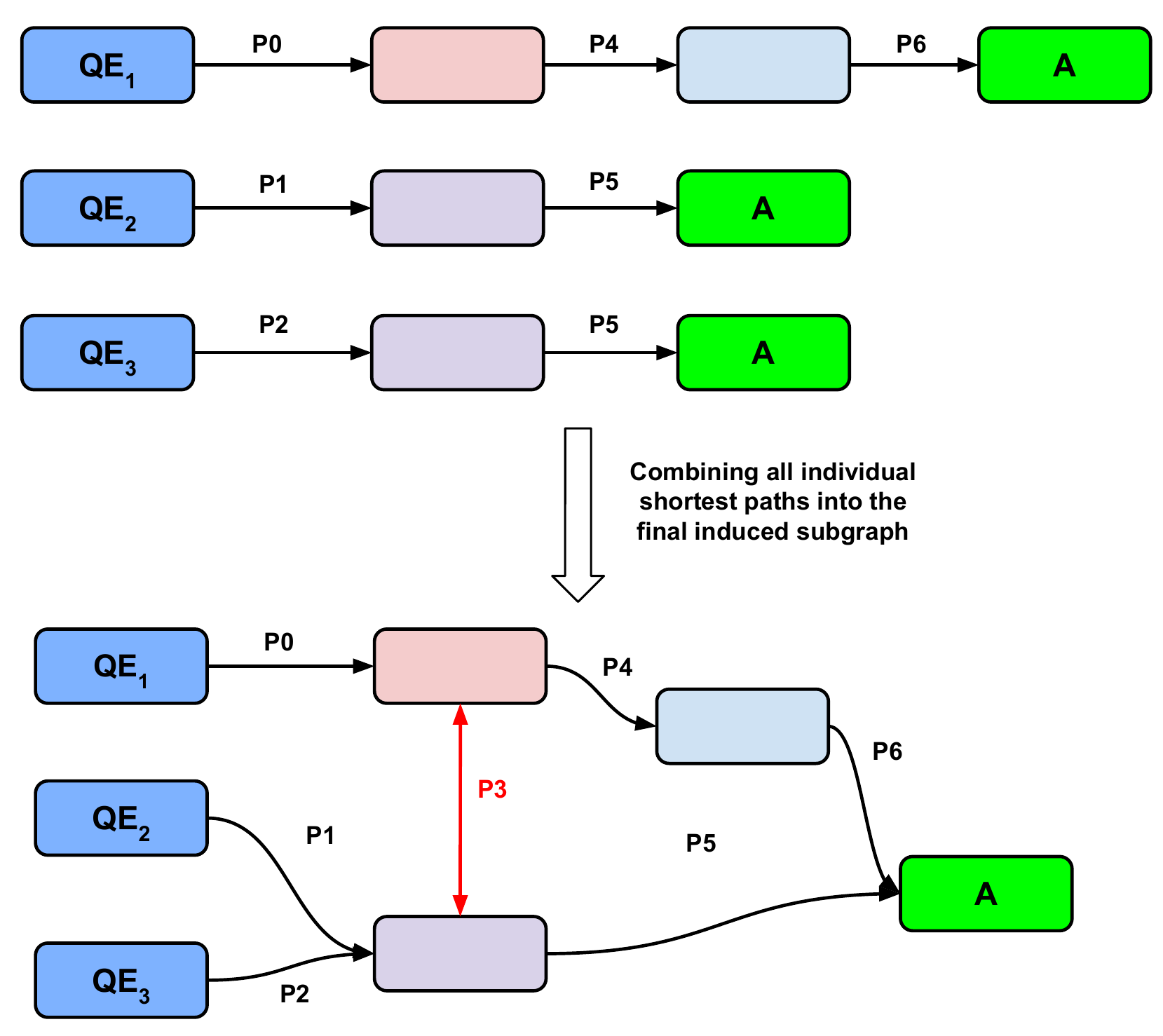}
  \caption{Extraction of subgraphs between question and answer in KG. We combine the extracted shortest paths into a single subgraph, adding links between intermediate nodes. Here, $\text{QE}_n$ is the Question Entities, $\text{A}$ is the current answer candidate entity, and the colored nodes represent the intermediate entities.}
  \label{fig:subgraph_construction_example}
\end{figure}
The first stage of our approach is subgraph extraction. We rely on the information conveyed in the relationships between question-answer pairs to construct subgraphs that capture the connectivity between question entities and candidate answers. For each candidate answer $a_i$, we employ a subgraph extraction algorithm that generates a KG's subgraph $G_i$ containing entities relevant to each question-answer pair and the shortest paths between them that contain relevant properties/relationships. This section presents the subgraph extraction algorithm and the process for constructing $G_i$ for each candidate.

Referring to our previous research, we utilize Wikidata to extract subgraphs representing the relationship between each question-answer pair. For this paper, we deploy a similar subgraph extraction protocol, further discussed in the following section, with various language models to generate answer candidates.

Note that subgraph extraction requires substantial computational resources; detailed hardware requirements for reproducing these experiments and all other details about used hardware are provided in Appendix~\ref{appx:hardware}.

\subsubsection{Answer Candidate Generation} \label{sec:base_model_expansion}
As the subgraph extraction protocol requires answer candidates, we need a source of distinct answer candidates for each question. Most LLM approaches for QA, such as the one presented by~\cite{DBLP:conf/coling/SenAS22-mintaka}, typically use Greed Search and evaluate the top-1 answer. However, it is important to note that the correct answer may not always be the top candidate. For example, the fine-tuned T5-XL-SSM~\cite{DBLP:conf/emnlp/RobertsRS20} model achieved higher Mean Reciprocal Rank (MRR) scores for our task, indicating that reranking could improve the {Hits@1} results. For an effective reranking pipeline, we require numerous unique answer candidates. However, even with Classical Beam Search, the output often consists of minor variations of a single sequence, which may not yield sufficient unique answer candidates for this reranking task.


To solve the problem, we apply Diverse Beam Search~\cite{vijayakumar2018diverse}, which produces a lot of candidates and generates them with higher variance. {Diverse Beam Search} is formulated as follows:
\begin{equation}
        Y_{[t]}^g = \underset{y_1^{g}, \dots, y_{B\prime}^g \in Y_t^g} {\text{argmax}} \quad \underbrace{\sum_{b \in [B\prime]} \Theta(y_{b, [t]}^g)}_{\text{diversity penalty}}  + \underbrace{\sum_{h=1}^{g-1} \lambda_g \Delta(y_{b,[t]}^g, Y_{[t]}^h)}_{\text{dissimilarity term}},
    \label{eq::quad_func}
\end{equation}

\noindent The formula involves splitting the set of beams at time $t$ into $g$ disjointed subsets $Y_{[t]}^g$, and then selecting the candidate with the highest diversity penalty, which is calculated as the sum of a diversity penalty function $\Theta(y_{b,[t]}^g)$ over all candidates in the subset. Additionally, a dissimilarity term is included, which is calculated as the sum of a dissimilarity function $\Delta(y_{b,[t]}^g, Y_{[t]}^h)$ over all previous subsets $Y_{[t]}^h$ up to time $g-1$. The dissimilarity term is weighted by a parameter $\lambda_g$. This formula is used to optimize the selection of answer candidates computationally efficiently.


We apply {Diverse Beam Search} to language models to generate diverse answer candidates. The specific models used, hyperparameters, and training details are described in Section~\ref{sec:base_model_training}.

\subsubsection{Question-Answer Subgraph Construction}
With our LM's produced answer candidates (Section~\ref{sec:base_model_expansion}) and the question entities, we construct the subgraph $G_i$ as the union of all shortest paths connecting each question entity with the candidate entity $a_i$. For each question-answer candidate pair, the desired subgraph $G_i$ is mathematically defined as an induced subgraph of the Wikidata KG $G$. Given our shortest paths from $e_i~\rightarrow~a_i$, where $e_i$ --- entity extracted from the question and $a_i$ --- candidate entity. For datasets where entities $e_i \in E$ is not annotated, we apply a lightweight Entity Linking strategy. We first prompt an LLM (GPT-OSS-120B\cite{openai2025gptoss120bgptoss20bmodel}) to extract entity surface forms from the question, then retrieve their Wikidata IDs via the Wikidata Search API, selecting the top result for each. 
We can use the following Algorithm~\ref{alg:sub_extract} to extract $G_i$. Let us define $H$ as the set of all distinct nodes within our shortest paths $P_i$. We want to preserve all edges between the nodes within $H$. We aim to retain the relationship between our question entities $E$ and candidate entity $a_i$ for all question-answer pairs. The process is schematically depicted in Figure~\ref{fig:subgraph_construction_example}. 


\begin{algorithm} 
\caption{Subgraph Extraction}\label{alg:sub_extract}
\begin{algorithmic}

\Require entities $E$, candidate $a_i$
\Ensure subgraph $G_i$

\For{$e_i$ \textbf{in} $E$}
\State shortest\_paths $\gets$ get\_shortest\_path\_from\_entity\_to\_candidate($e_i$, $a_i$)
\EndFor
\State $H$ $\gets$ get\_unique\_nodes\_shortest\_paths\_flattened(shortest\_paths)
\State $G_i$ $\gets$ DirectedGraph()

\For{unique\_node \textbf{in} $H$}
\State unique\_node\_neighbor $\gets$ get\_neighboring\_nodes(unique\_node)

    \For{neighbour\_node \textbf{in} unique\_node\_neighbor}
        \If{neighbour\_node \textbf{in} $H$}
            \State $G_i$.add\_edge\_between(unique\_node, neighbour\_node)
        \EndIf   
    \EndFor
\EndFor

\noindent \Return $G_i$
\end{algorithmic}
\end{algorithm} 
The specific implementation details for constructing subgraphs, including the tools and data sources used, are described in Section~\ref{sec:implementation_details}.

\subsection{Feature Engineering} \label{sec:subgraph_features}
After extracting subgraphs $G_i$ for all answer candidates of our LMs, we use all possible useful features for reranking. Referring to our previous study, we mainly focused on a simple text representation of the extracted subgraphs to rank our answer candidates. Thus, in this study, we propose extracting as many useful features as possible and analyzing each feature's importance in this reranking problem. We have divided the features into the following main categories: graph, text, and Graph2Text sequence features. 

\subsubsection{Graph Features} \label{sec:graph_features}
With our extracted subgraphs and their corresponding answer candidate, we seek to use the relationship from the subgraphs to classify the correct answer candidate. As the first simple baseline, we utilize graph features consisting of simple numerical subgraph statistics. We hypothesize that subgraphs with the correct answer will be less ``complex'' than subgraphs with the incorrect answer candidate. Therefore, we would want the graph features to convey the complexity of the respective subgraph. With a clear objective in mind, we experiment with the following graph features:  

\begin{itemize}
    \item \textbf{Number of nodes and edges}: basic statistics of the nodes and edges of graph $G$.
    \item \textbf{Number of cycles}: a cycle of graph $G$ is a non-empty path that starts from a given node and ends at the same node. 
    \item \textbf{Number of bridges}: a bridge of graph $G$ is an edge, where its deletion increases the number of connection components. 
    \item \textbf{Average shortest path}: the average of each shortest path between the question entity and the answer entity. 
    \item \textbf{Density}: measurement of the density of a graph, where the number of edges in a dense graph is close to the maximal number of edges (each pair of nodes is connected by an edge). The density $d$ for the graph $G$ is formulated as $d = \frac{m}{n(n-1)}$, where $n$ is the number of nodes and $m$ is the number of edges in $G$.
   \item \textbf{Katz centrality}~\cite{katz1953new}: measurement of the importance (or ``centrality'' - how ``central'' a node is in the graph) of a specific node $i$ in a graph $G$. The Katz centrality for node $i$ of graph $G$ is formulated as $x_i = \alpha \sum_{j} A_{ij} x_j + \beta$, where $A$ is the adjacency matrix of graph $G$ with eigenvalues $\lambda$, $\beta$ is the parameter that controls the initial centrality, and $\alpha < \frac{1}{\lambda_{\max}}$. 
    \item \textbf{PageRank}~\cite{page1999pagerank}: a popular algorithm used by Google to rank web pages in the search query by counting the number and quality of links to a page to determine an estimate of its importance. In graph theory, the ``web pages'' and ``links'' are synonymous with nodes and edges. 
\end{itemize}

\noindent We hypothesize that these features may provide ranker models with insights into the complexity of the respective subgraphs.

\subsubsection{Text Features} \label{sec:text_features} As researched in our original paper~\cite{DBLP:conf/paclic/SalnikovLRNBMP23-originalpaper}, the ablation study showcased the importance of including the question within the text representation of the subgraph. Therefore, besides the simple graph features, we want to emphasize each question/answer pair without using extracted subgraphs. Thus, the text features represent the concatenation between the question and answer, separated by a semicolon --- ``\texttt{;}''. To use this simple concatenation for all ranker approaches, we encode the string using the MPNet\footnote{\url{https://huggingface.co/sentence-transformers/all-mpnet-base-v2}} embedding model~\cite{DBLP:conf/nips/Song0QLL20}, discussed more in~\ref{appx:detailed_results}.

\subsubsection{Graph2Text Sequence Features} \label{sec:g2t_seq} 

Given the vast amount of data contained in KGs, it is essential to convert this information into natural language to facilitate understanding and accessibility. Converting a KG into text, known as KG-to-text or Graph2Text, has demonstrated notable success in various applications~\cite{DBLP:journals/corr/abs-2309-11206}. Therefore, when generating text from a KG, it is crucial to analyze the underlying graph structure carefully to ensure accurate translation.

Without an obvious way of incorporating the question within the subgraphs, relying purely on the subgraphs to rerank is ineffective~\cite{DBLP:conf/paclic/SalnikovLRNBMP23-originalpaper}. Therefore, we address this issue by further exploration of different KG-to-text methods. The main objective is experimenting with various techniques to represent the extracted subgraphs more explicitly. For this type of textual feature, we researched and developed three methods for representing subgraphs as a text, including \textbf{Graph2Text Deterministic,  Graph2Text T5, and  Graph2Text GAP}.

Firstly, we employ the \textbf{Graph2Text Deterministic} approach, the most straightforward text linearization approach. In simple terms, the subgraphs are unraveled by their matrix representation. Firstly, to linearize, we convert the subgraph into its binary adjacency matrix representation, $A$. 
Given $n$ nodes in the subgraph, the resulting matrix's dimension will be $n \times n$. The matrix's element $[i, j]$ represents the existence of an edge between a node with index $i$ and a node with index $j$. Then, we replace the edges in the matrix with the edge label and call it $A'$. 
Lastly, we unravel $A'$ row by row to produce our final sequence and add the triple (node\_from, edge, node\_to) to our final sequence. Algorithm \ref{alg:sub2seq} summarizes the aforementioned steps. 

\begin{algorithm}
\caption{Subgraphs to Sequence}\label{alg:sub2seq}
\begin{algorithmic}
\Require Subgraph G
\Ensure Text representation of subgraph Seq

\State adj\_matrix $\gets$ get\_adjacency\_matrix(G)
\State Seq $\gets$ ""
\For{i \textbf{in} adj\_matrix}
    \For{j \textbf{in} i}
        \If{j \textbf{not} 0}
            \State edge\_info = get\_edge\_between\_nodes(G, $i$, $j$)
            \State Seq $\gets$ Node($i$).label + edge\_info + Node($j$).label
        \EndIf
    \EndFor
\EndFor

\noindent \Return Seq
\end{algorithmic}
\end{algorithm}

\begin{table}
\centering
\caption{Statistics of the WebNLG 2.0 parallel knowledge graph-to-text dataset.}
\label{tab:webnlg_label}

\begin{subtable}[t]{0.48\textwidth}
\centering
\begin{tabular}{lr}
    \toprule
    \textbf{Entities} & 2,730 \\
    \textbf{Relations} & 354 \\
    \textbf{Triples} & 81,927 \\
    \bottomrule
\end{tabular}
\caption{Knowledge Graph statistics. Total number of KG components, number of tokens in the narratives.}
\label{tab:kg_stats}
\end{subtable}
\hfill
\begin{subtable}[t]{0.48\textwidth}
\centering
\begin{tabular}{lr}
    \toprule
    \textbf{Total} & 623,902 \\
    \textbf{Unique} & 8,075 \\
    \textbf{Entity} & 60\% \\
    \bottomrule
\end{tabular}
\caption{Texts statistics. The percentage of text entities represents the portion of the text that includes entity labels.}
\label{tab:texts_stats}
\end{subtable}



\end{table}

For the remaining two text linearization approaches, \textbf{Graph2Text T5} and \textbf{Graph2Text GAP}, we employ more complex neural-based models trained on the WebNLG~2.0 dataset~\cite{DBLP:conf/acl/GardentSNP17}. This dataset consists of instances, where each includes a KG from DBpedia~\cite{dbpedia} and a target text comprising one or more sentences that describe the graph. The test set is divided into partitions of seen (DBpedia categories present in the training set) and unseen (DBpedia categories not present in the training set). The statistics of this hand-crafted and human-verified dataset are described in detail in Table~\ref{tab:webnlg_label}. 

The idea behind the \textbf{Graph2Text T5} approach is to extract informative and useful features from KGs using pre-trained text-to-text LMs. With the impressive capabilities of pretrained LMs in the text-to-text generation task, we seek to replicate such results in the graph-to-text scope. Our idea is built upon the analogous algorithm discussed in~\cite{DBLP:conf/acl/investigatingPretrained}. The authors tackle the graph-to-text generation task in this work with two popular text-to-text pre-trained LMs, BART and T5. These models have an encoder-decoder architecture, which makes them well-suited for conditional text generation tasks. To adapt these models for the graph-to-text task, the authors continue pre-training BART and T5 using the following approaches:
\begin{enumerate}
    \item Language Model Adaptation (LMA): the models are trained on reference texts that describe graphs, following the BART and T5 pre-training strategies.
    \item Supervised Task Adaptation (STA): the models are trained on pairs of graphs and their corresponding texts collected from the same or a similar domain as the target task --- graph-to-text in this case. 
\end{enumerate}

Building on the STA approach via T5 and WebNLG~2.0, we obtain graph-to-text sequences by first converting the graph into a sequence of tokens through linearization. We use the string ``convert the [graph] to [text]:'' to acquire this linearised sequence. This output sequence is then fed into the input sequence for the T5 model tuned on WebNLG~2.0. The specific training hyperparameters for the Graph2Text T5 approach are detailed in Section~\ref{sec:g2t_training}.

Lastly, the \textbf{Graph2Text GAP} approach is based on the current state-of-the-art graph-to-text task, GAP, built on BART~\cite{DBLP:conf/coling/ColasAW22-GAP}. 
The main idea of GAP is a fully graph-aware encoding combined with the coverage of pre-trained LMs. The GAP KG-to-text framework fuses graph-aware elements into existing pre-trained LMs, capturing the advantages brought forth by both model types. The architecture of this solution consists of two main components:

\begin{enumerate}
\item \textbf{Global Attention}: to capture the graph's global semantic information, the graph’s components are first encoded using an LM. This allows the model to leverage the lexical coverage of pre-trained LMs.
\item \textbf{Graph-aware Attention}: to attend to and update the representations of entities, relations, or both, a topological-aware graph attention mechanism was introduced, which includes entity and relation type encoding.
\end{enumerate}

Applying the work of GAP, we first linearize the input graph into a text string by creating a sequence of all triples in the KG, interleaved with tokens that separate each triple and the triple's components (head, relation, and tail). Then, we use a transformer encoder to obtain vector representations. The first module in each transformer layer acts as a Global Attention and captures the semantic relationships between all tokens. Moreover, we use a Graph-aware Attention module to capture the sparse nature of adjustment in a graph and apply it to entity and relation vectors from word vectors. By proposing this flexible framework, where graph-aware components can be interchanged, the current architecture aims to generate coherent and representative text descriptions of the KG. Like the Graph2Text T5 approach, we pretrain the model on the WebNLG~2.0 dataset and get the final predictions through the fine-tuned model. The specific training hyperparameters for the Graph2Text GAP approach are detailed in Section~\ref{sec:g2t_training}.

In this research, we introduce the more complex neural-based graph-to-text approach to explore further the reranking capabilities of the textual representation of our extracted subgraphs. The initial rudimentary text linearization approach has already achieved state-of-the-art Hits@1. We look for a more complete case study on reranking the text linearization with these two neural-based linearization approaches. To further digest the two methods, a comparison between the Graph2Text T5 and Graph2Text GAP sequences can be seen in the table ~\ref{tab:example_of_predictions_label}. Additionally, for better visualization, we implement a web application that automatically applies the T5 and GAP approaches to the desired subgraph, discussed in detail in Section ~\ref{sec:webapp}.

\begin{table}[!htb]
    \caption{Examples of subgraphs and their corresponding generated texts for various feature extractors.}
    \label{tab:example_of_predictions_label}
    \begin{tabular*}{\textwidth}{m{0.42\textwidth}|p{0.25\textwidth}|p{0.25\textwidth}}
        \toprule
        \multicolumn{1}{c}{Subgraph} & \multicolumn{2}{c}{Corresponding Graph2Text} \\
         & \multicolumn{1}{c}{T5} & \multicolumn{1}{c}{GAP} \\
        \midrule \\
        \includegraphics[width=0.28\textwidth]{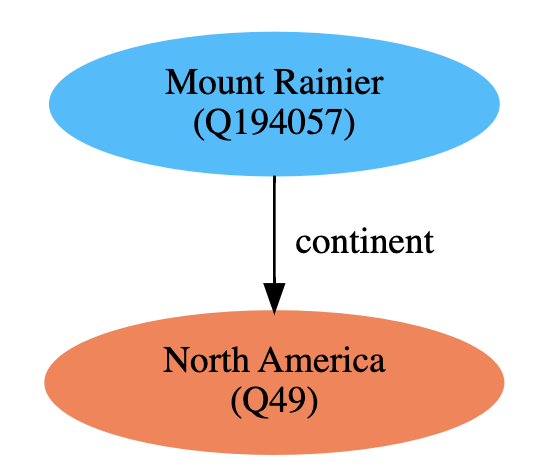} & Mount Rainier is located in North America. & The summit of North America is called Mount Rainier. \\ \hline

        \includegraphics[width=0.41\textwidth]{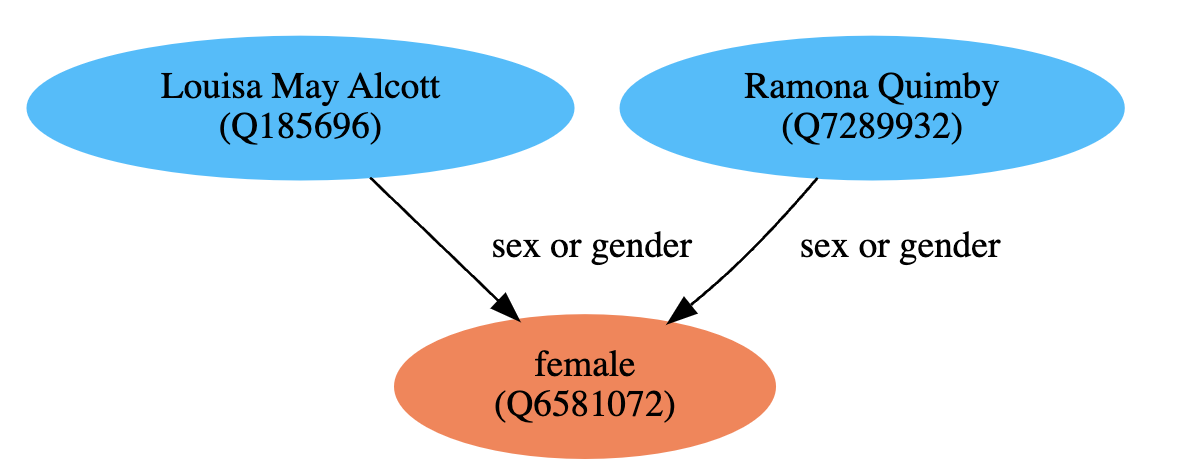} & Louisa May Alcott and Ramona Quimby are both females. & Louisa May Alcott and Ramona Quimby are both men. \\ \hline

        \includegraphics[width=0.41\textwidth]{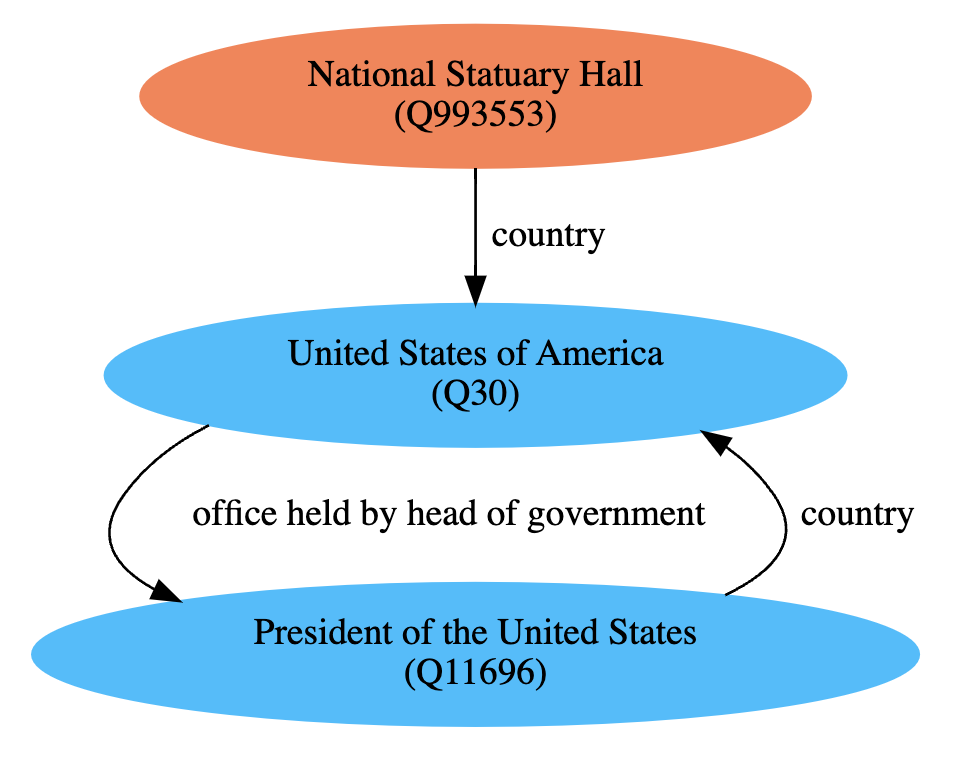} & The United States of America is the location of the National Statuary Hall and the office of the President of the United States. & The National Statuary Hall is located in the United States of America, where the leader is known as the President. The country is the location of the US congress and has the President as its President. \\ \hline

        \includegraphics[width=0.41\textwidth]{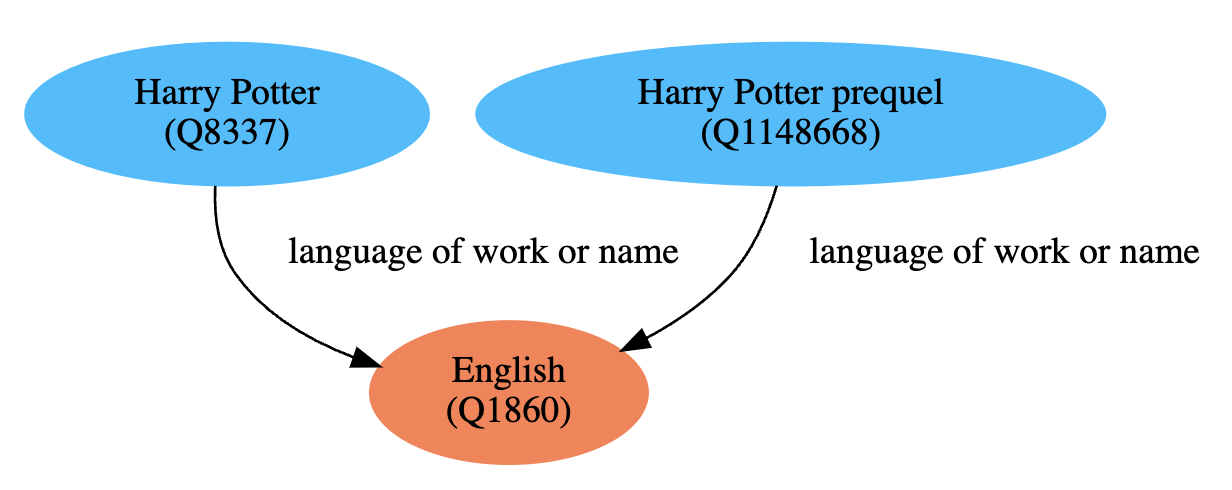} & The Harry Potter prequel is written in English, as is the Harry Potter book. & English is the language of both Harry Potter prequel and the English book. \\
        
        \hline
    \end{tabular*} 
\end{table}

All three variation of Graph2Text Sequence features are further encoded with MPNet embedding model~\cite{DBLP:conf/nips/Song0QLL20}, discussed more in \ref{appx:detailed_results}. Moreover, motivated by our previous research, we employ context and highlight these Graph2Text sequences, discussed further in \ref{hl_context}.

\subsection{Rankers} \label{ranker}
With the subgraphs and their extracted features discussed above, we devise several reranking approaches to maximize the performance of the base models. As the focal point of the research is the reranking scope, we employ reranking methodologies from least to most complex. The hypothesis is a positive trend in performance as we apply more complex models and features.

As a starting point, we employ semantic reranking. This is a popular solution in information retrieval~\cite{DBLP:journals/jifs/FigueroaPP20, DBLP:journals/corr/HendersonASSLGK17}, implemented differently under the same name. Building on this foundation, our semantic ranker utilizes the MPNET ~\cite{DBLP:conf/nips/Song0QLL20} embeddings of the answer candidates, further justified in~\ref{mpnet_explain}. We then rank the answer candidates by the cosine similarity between the embedding vectors. 

To establish a robust baseline, we use a Majority Vote strategy, also known as self-Consistency~\cite{wang2022self}. This method takes advantage of the redundancy in the beam search output by selecting the candidate answer that appears most frequently among the generated sequences. In contrast to our proposed methods, this baseline does not use any external knowledge from knowledge graphs or subgraphs. 

In the next layer of complexity, we utilize regression-based models, namely, linear and logistic regression. 
For the features set, we apply all features discussed in~\ref{sec:subgraph_features} (for features in text/string format, we apply MPNET embeddings, discussed in~\ref{mpnet_explain}). In the case of linear regression, we employ ordinary least squares linear regression to predict either $1$ or $0$, corresponding to correct and incorrect responses, respectively. The predicted score was then used to rank the potential answers by sorting the values from highest to lowest. Although we employ logistic regression for the same reranking task, we reformat the problem to a classic classification problem. We sort the answers with the highest classification confidence to rank the candidates. We use a standard logistic regression model with L2 regularisation. 

In addition to regression-based models, we want to utilize the same features with a more complex ranker. Thus, we explore and experiment with the gradient boosting model, specifically the CatBoost regression model~\cite{DBLP:conf/nips/ProkhorenkovaGV18-catboost}. The specific hyperparameters and training details for the CatBoost ranker are described in Section~\ref{sec:ranker_training}. Like linear regression, we use the predicted score to rank the answer candidates by sorting the values.

Last but not least, the most effective and complex approach for reranking answer candidates is a neural-based ranker with textual features as input. We experiment with a transformer-based model with an additional regression head layer. The specific training hyperparameters and optimization details for the neural ranker are described in Section~\ref{sec:ranker_training}. To keep the experiments clear and transparent, we keep the same MPNet model for this ranker. We employ this variation of sentence transformer throughout this research for various sentence/text embeddings, as mentioned in~\ref{mpnet_explain}. Due to the lack of a straightforward method for utilizing numeric or table-like features with this ranker, we choose not to apply graph features.

\section{Experiments}
In this section, we describe an experimental setup to evaluate the effectiveness of our proposed approach for reranking LLM-generated candidate answers using Knowledge Graph subgraphs. We explore the impact of different combinations of features (graph features and textual representations) on reranking performance and examine the effects of various learning-to-rank models, ranging from simple to more sophisticated, on reranking Hits@1 accuracy.

\subsection{Dataset} \label{sec:dataset}
To further enhance the results gathered in the original paper, we also conduct our research on Mintaka~\cite{DBLP:conf/coling/SenAS22-mintaka} dataset, which is a large-scale, complex and natural dataset that can be used for end-to-end question-answering models, composed of $20,000$ question-answer pairs. This dataset is annotated with Wikidata entities and comprises 8 types of complex questions. These types include: 
\begin{itemize}
    \item \textbf{Count}, e.g., Q: How many astronauts have been elected to Congress? A: 4.
    \item \textbf{Comparative}, e.g., Q: Is Mont Blanc taller than Mount Rainier? A: Yes.
    \item \textbf{Superlative}, e.g., Q: Who was the youngest tribute in the Hunger Games? A: Rue.
    \item \textbf{Ordinal}, e.g., Q: Who was the last Ptolemaic ruler of Egypt? A: Cleopatra.
    \item \textbf{Multi-hop}, e.g., Q: Who was the quarterback of the team that won Super Bowl 50? A: Peyton Manning.
    \item \textbf{Intersection}, e.g., Q: Which movie was directed by Denis Villeneuve and stars Timothee Chalamet? A: Dune.
    \item \textbf{Difference}, e.g., Q: Which Mario Kart game did Yoshi not appear in? A: Mario Kart Live: Home Circuit.
    \item \textbf{Yes/No}, e.g., Q: Has Lady Gaga ever made a song with Ariana Grande? A: Yes.
    \item \textbf{Generic}, e.g., Q: Where was Michael Phelps born? A: Baltimore, Maryland.
\end{itemize}
This research centers around the reranking aspect of the pipeline, discussed in our previous research~\cite{DBLP:conf/paclic/SalnikovLRNBMP23-originalpaper}. Thus, with the question types listed above, we exclude {Yes/No} and {Count} questions. These question types offer no value information, as numbers and ``yes/no'' have no respective Wikidata entities, leading to a non-existent/impractical relationship between the question/answer pair. Thus, we deem {Yes/No} and {Count} pointless in the scope of our research. However, one could still utilize the entire pipeline to compute and evaluate the results based on the whole Mintaka dataset, including {Yes/No} and {Count}. Referring to the question type classifier introduced in our original research ~\cite{DBLP:conf/paclic/SalnikovLRNBMP23-originalpaper}, this additional component allows for {Yes/No} and {Count} questions to receive special treatment.

We also compile and publish\footnote{\url{https://huggingface.co/datasets/s-nlp/KGQASubgraphsRanking}} the dataset of subgraphs for the whole Mintaka dataset (for train, validation, and test splits separately). Additionally, we also publish the answer candidates generated by the base LMs. Subgraphs are collected using the process discussed in Section~\ref{sec:subgraph_extract}: we generate candidate answers, we take the true answer and the entities from the question entity neighbors as candidates, and construct subgraphs with Algorithm~\ref{alg:sub_extract}. As a result, we construct a ``correct'' subgraph containing the correct highlighted answer and several ``incorrect'' subgraphs from the incorrect candidate answers generated by the model. 
We present four versions of the dataset with subgraphs: with candidates generated by {T5-Large-SSM}, {T5-XL-SSM}, {Mistral}, and {Mixtral} models.

In addition to Mintaka, we conduct additional experiments on the MKQA dataset\cite{mkqa}, an open-domain question answering evaluation set comprising 10,000 question-answer pairs aligned across 26 typologically diverse languages, for consistency, we are used English QA pairs. The dataset contains queries sampled from the Google Natural Questions dataset with passage-independent answers annotated with Wikidata QIDs, enabling entity linking and knowledge graph-based evaluation. The specific heuristic used to map MKQA answers to Wikidata nodes is detailed in Appendix~\ref{appx:mkqa_linking}. We prepare subgraphs for MKQA using the same process as for Mintaka, following the subgraph extraction protocol described in Section~\ref{sec:subgraph_extract}. For MKQA, we generate subgraphs only for answer candidates produced by {T5-Large-SSM} and {T5-XL-SSM} models.

\subsection{Base Model Training} \label{sec:base_model_training}
We apply Diverse Beam Search to generate answer candidates using the following language models: {T5-large-ssm}~\cite{DBLP:conf/emnlp/RobertsRS20}, {T5-XL-ssm}~\cite{DBLP:conf/emnlp/RobertsRS20}, {Mistral}~\cite{DBLP:journals/corr/abs-2310-06825-mistral}, and {Mixtral}~\cite{DBLP:journals/corr/abs-2401-04088-mixtral}. We extend our previous research~\cite{DBLP:conf/paclic/SalnikovLRNBMP23-originalpaper} by fine-tuning the proposed T5-like models and comparing them to more recent state-of-the-art models like Mistral and Mixtral, which should make our research more applicable to real-world use cases. T5-large-SSM and T5-XL-SSM were reported to be state-of-the-art both in the original Mintaka paper and our previous work, serving as a good baseline for comparison in this study.

For Diverse Beam Search, we use the following hyperparameters: $200$ beams, $20$ beam groups, and a $0.1$ diversity penalty. To finetune the T5-like models, we first train them on English questions for $10000$ steps, following the protocols outlined in the original Mintaka paper~\cite{DBLP:conf/coling/SenAS22-mintaka}. For the more state-of-the-art Mistral and Mixtral, we finetune with LoRA and train on English questions by generating the answer candidates with ``\textit{Answer as briefly as possible without additional information. [Question]}''. However, for the T5-like models, despite adhering to these protocols, we could not achieve the reported Hits@1 accuracy in the original paper. Despite this challenge, the main focus of the study is on the reranking aspect of the pipeline. Therefore, this paper's primary contribution is improving our fine-tuned models.

\subsection{Graph2Text Model Training} \label{sec:g2t_training}
For the Graph2Text T5 approach, we use the following hyperparameters: \textit{learning rate}: $1e^{-3}$, \textit{batch size}: 4, \textit{gradient accumulation steps}: 32, and \textit{Adam optimizer}. The model is trained on the WebNLG~2.0 dataset~\cite{DBLP:conf/acl/GardentSNP17}, which consists of instances where each includes a KG from DBpedia~\cite{dbpedia} and a target text comprising one or more sentences that describe the graph. The test set is divided into partitions of seen (DBpedia categories present in the training set) and unseen (DBpedia categories not present in the training set). The statistics of this hand-crafted and human-verified dataset are described in detail in Table~\ref{tab:webnlg_label}.

For finetuning the Graph2Text GAP approach, we use the following hyperparameters: \textit{learning rate}: $2e^{-5}$, \textit{batch size}: 16; \textit{beam size}: 5, \textit{Adam Optimizer}, 50 \textit{nodes}, and 60 \textit{relations}. Like the Graph2Text T5 approach, we pretrain the model on the WebNLG~2.0 dataset and get the final predictions through the fine-tuned model.

\subsection{Ranker Training} \label{sec:ranker_training}
For the CatBoost regression model~\cite{DBLP:conf/nips/ProkhorenkovaGV18-catboost}, we use grid search to finetune \textit{learning\_rate, depth,} and \textit{iteration} hyperparameters. We use root-mean-square deviation (RMSE) to evaluate during the grid search process.

For the neural-based ranker with textual features as input, we use a transformer-based model with an additional regression head layer, which is fine-tuned using {mean-square loss} and {AdamW} optimisation~\cite{DBLP:conf/iclr/LoshchilovH19-adamw}.

\subsection{Implementation Details} \label{sec:implementation_details}
Given the computational limitations of Wikidata Query Services, we cannot extract the shortest paths to construct each respective subgraph using the subgraph extraction algorithm. A time-out protocol exists for each SPARQL shortest path query to Wikidata Query Services. As a solution to this limitation, we utilize \textit{igraph}\footnote{\url{https://igraph.org}}, a library consisting of network analysis tools with an emphasis on efficiency and portability. With \textit{igraph}, we parse the entire Wikidata KG via Wikidata's online RDF dumps\footnote{\url{https://dumps.wikimedia.org/wikidatawiki/entities/}}. By building the parsed local Wikidata KG via \textit{igraph}, we can efficiently construct the subgraphs dataset with the extraction algorithm discussed in Algorithm~\ref{alg:sub_extract}.

\subsection{Question Entities} \label{ques_ent}
Referencing the subgraph extraction protocol discussed in~\ref{sec:subgraph_extract}, we require question entities and the answer entity for each question-answer pair. Regarding the question entities, any entity linker such as {mGENRE}~\cite{decao2021multilingual} could be applied. However, with the objectives outlined above, utilizing an entity linker would derail the main focus of evaluating this reranking scope. As a result, we leverage the golden truth question entities provided by the Mintaka dataset. 
\subsection{Text Embeddings} \label{mpnet_explain}
Some features derived from the extracted subgraphs are in their natural language form (answer candidates for semantic ranker, Graph2Text sequences, and text features). Therefore, we use these features for our reranking objectives based on the MPNet embedding model~\cite{DBLP:conf/nips/Song0QLL20}. In Performance Sentence Embeddings (evaluation of the quality of the embedded sentence) and Performance Semantic Search (evaluation of the quality of the embedded search queries \& paragraph), the MPNet Embedding model outperforms 37 other sentence transformer models~\cite{reimers-2019-sentence-bert}. These models were compared by averaging the Performance Sentence Embeddings and Performance Semantic Search while considering the speed and the model size. In addition to the highest embedding performance, MPNet is relatively small while fast in training time. Moreover, this model is very popular within the HuggingFace community \footnote{\url{https://huggingface.co/sentence-transformers/all-mpnet-base-v2}}. Utilizing a well-known and widespread embedding model would enhance the aim of formulating our approach as a reranking problem motivated by end-user requirements. 

\subsection{Graph2Text with Highlight \& Context} \label{hl_context}
As mentioned in~\ref{sec:g2t_seq}, we focus on different text representations of the subgraphs to rank the respective answer candidates. 
We employ context for linearised text representation of the subgraph. This addition is a simple concatenation between the question and the linearised sequence, separated by a special token $<$/s$>$ to emphasize the question in the question-answer pairing. Moreover, the study showcased the effectiveness of highlighting~(HL) the answer candidate within the linearised sequence of the concatenation. 
The context is motivated by the assumption that the subgraph alone does not provide the necessary information to answer the question. In other words, the model cannot answer the question without it. Similarly, it is difficult to rank the answers if the model has no idea which entities in the subgraph are potential answer candidates.
An example of such concatenation, the rank by MPNet to achieve the current state-of-the-art, is shown below:  

\textbf{Question}: Which actor was the star of Titanic and was born in Los Angeles, California?

\textbf{Answer}: Leonardo DiCaprio
\vskip 0.15in
\textbf{Original Deterministic Sequence with HL and Context\cite{DBLP:conf/paclic/SalnikovLRNBMP23-originalpaper}}:

Which actor was the star of Titanic and was born in Los Angeles, California? $<$/s$>$ [unused1]Leonardo DiCaprio[unused2], place of birth, Los Angeles, Titanic, cast member, [unused1]Leonardo DiCaprio[unused2]

\noindent We employ a similar \textbf{HL} and \textbf{context} protocol to our three Graph2Text sequences. Similar to the original approach~\cite{DBLP:conf/paclic/SalnikovLRNBMP23-originalpaper}, the objective is to assist the model in understanding the question-answer pair and the Graph2Text sequence. An example of such processing for the three sequences can be seen below:

\textbf{Question}: Which actor was the star of Titanic and was born in Los Angeles, California?

\textbf{Answer}: Leonardo DiCaprio
\vskip 0.15in
\textbf{Graph2Text Deterministic}: Which actor was the star of Titanic and was born in Los Angeles, California? $<$/s$>$ [unused1]Leonardo DiCaprio[unused2], place of birth, Los Angeles, Titanic, cast member, [unused1]Leonardo DiCaprio[unused2]
\vskip 0.1in
\textbf{Graph2Text T5}: Which actor was the star of Titanic and was born in Los Angeles, California?$<$/s$>$ Los Angeles born [unused1]Leonardo DiCaprio [unused2], who played the role of Jack Sparrow in the film Titanic, was born in the United States.
\vskip 0.1in
\textbf{Graph2Text GAP}: Which actor was the star of Titanic and was born in Los Angeles, California?$<$/s$>$Born in Los Angeles, the actor, [unused1]Leonardo DiCaprio [unused2], was a member of the crew of the Titanic.

\noindent It is important to note that we employ the \textbf{context} and \textbf{HL} approaches only for the MPNet approach, discussed in \ref{ranker}. Sensibly, the transformer-based model trained on sentences and paragraphs is adept at extracting useful information from natural texts. Thus, unlike our other proposed approaches, the MPNet approach classically only handles text embedding as input. Therefore, we utilize \textbf{context} and \textbf{HL} to give the model the best chance at extracting useful information to rank answer candidates. Other approaches, such as regression-based and gradient boosting, have various other features (i.e., graph, text, and sequence). Thus, we choose not to employ \textbf{context} and \textbf{HL} transformation for Graph2Text sequences for regression-based and gradient-boosting rankers. We discuss the pipeline for each ranker in more detail in~\ref{experimental_pipeline}.

\subsection{Experimental Pipeline} \label{experimental_pipeline}
With our two objectives of observing the effects of 1) different combinations of feature sets and 2) different reranking approaches in varying complexity, we devise our experiments for each answer candidate source (from either T5-Large-SSM, T5-XL-SSM, Mistral, or Mixtral)  as the following: 
\begin{itemize}
    \item apply each feature set extracted from the subgraphs (from the least to most complex) to each proposed ranker (from the least to most complex). The feature sets $A, B, C$ are ranked from least to most complex. We feed $A$, then $B$, then $C$ to each ranker. The goal is to observe the performance of varying complexity feature sets with varying complexity rankers. 
    \item add/combine different feature sets (from the least to most complex) to each proposed ranker (from the least to most complex). The feature sets $A, B, C$ are ranked from least to most complex. We feed $A$, then $A+B$, then $A+B+C$ to each ranker. The goal is to observe how adding more complex feature sets affects the performance of each ranker. 
\end{itemize}
With the above experimental pipeline, we seek to provide an exhaustive case study on reranking LLM's answer candidates with rankers and feature sets of varying complexity.

\subsection{Evaluation} \label{evaluation}
As outlined in~\ref{experimental_pipeline}, we experiment with numerous feature sets and rankers. Leveraging the limited amount of answer candidates, the primary objective is to determine whether selecting a final answer from this set is appropriate for our question-answering task. We evaluate using the {Hits@N} metric for all experiments discussed. Even if the top answer (Hits@1) is incorrect, this metric {Hits@N} allows us to understand the potential effectiveness of the respective ranker. For example, let us define two QA systems that can provide dozens of possible answers. These two systems generate the correct answer at the second and tenth positions. From an end-user point of view, the system that provides the correct answer at the second position or earlier in the sequence of answers will be much more beneficial. With that being said, as our research's main focus is the reranking task, Hits@N will provide us with valuable insights into each feature set and ranker.   
 
\section{Results \& Discussion}

\begin{figure}
    \centering
    \includegraphics[width=0.95\linewidth]{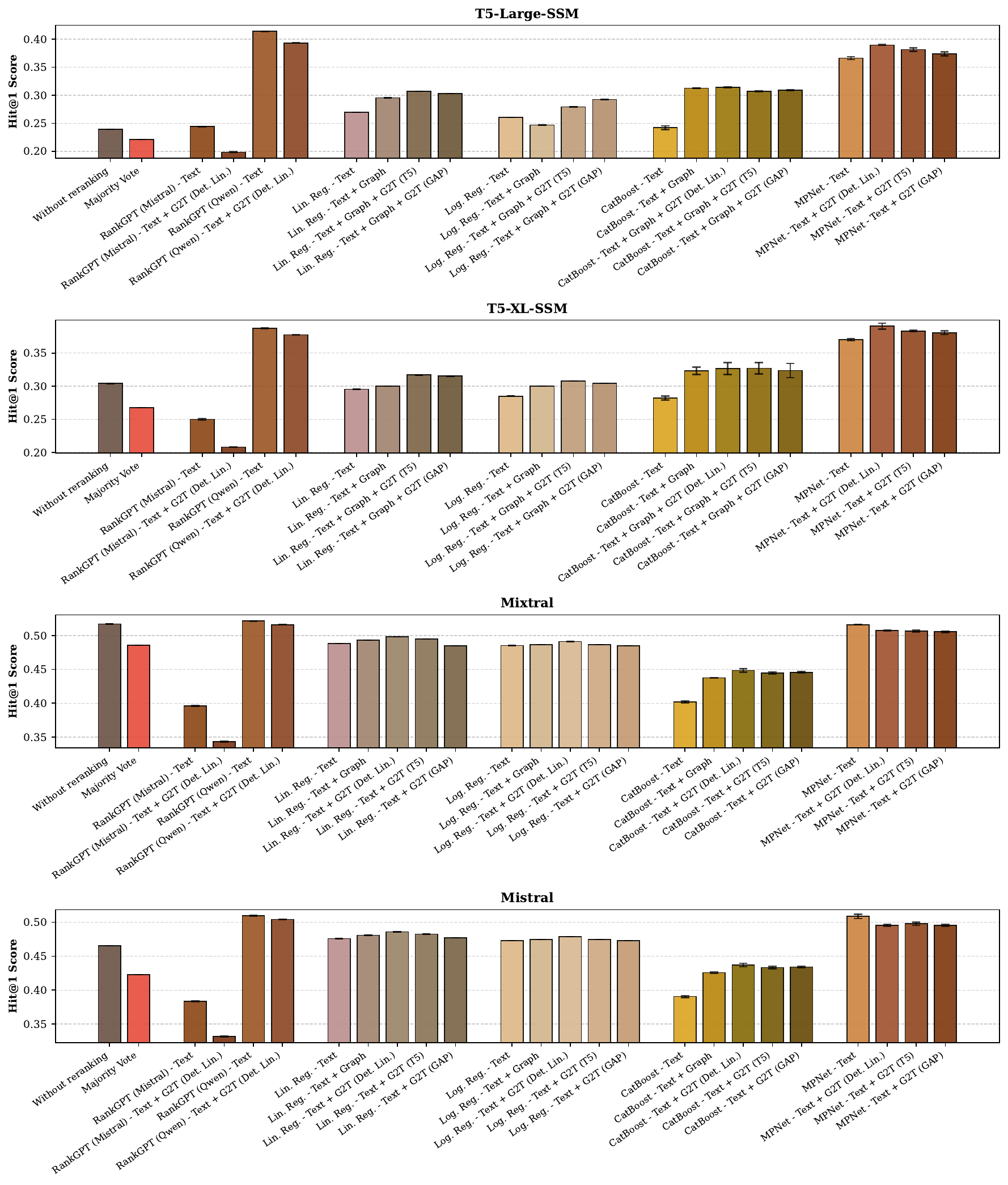}
    \caption{Comparative analysis of reranking methods applied to candidate pools generated by T5-Large-SSM, T5-XL-SSM, Mistral-7B, and Mixtral-8x7B. Bars represent the Hit@1 score of the top-ranked answer on \textbf{Mintaka} dataset. Methods are grouped by model architecture: Baseline (Without reranking, Majority Vote), RankGPT, Regression Baselines (Linear/Logistic), Gradient Boosting (CatBoost), and proposed Neural Ranker (MPNet). Error bars indicate standard deviation over 3 runs for non-deterministic methods.}
    \label{fig:main_results_mintaka_bars}
\end{figure}

\begin{figure}
    \centering
    \includegraphics[width=0.95\linewidth]{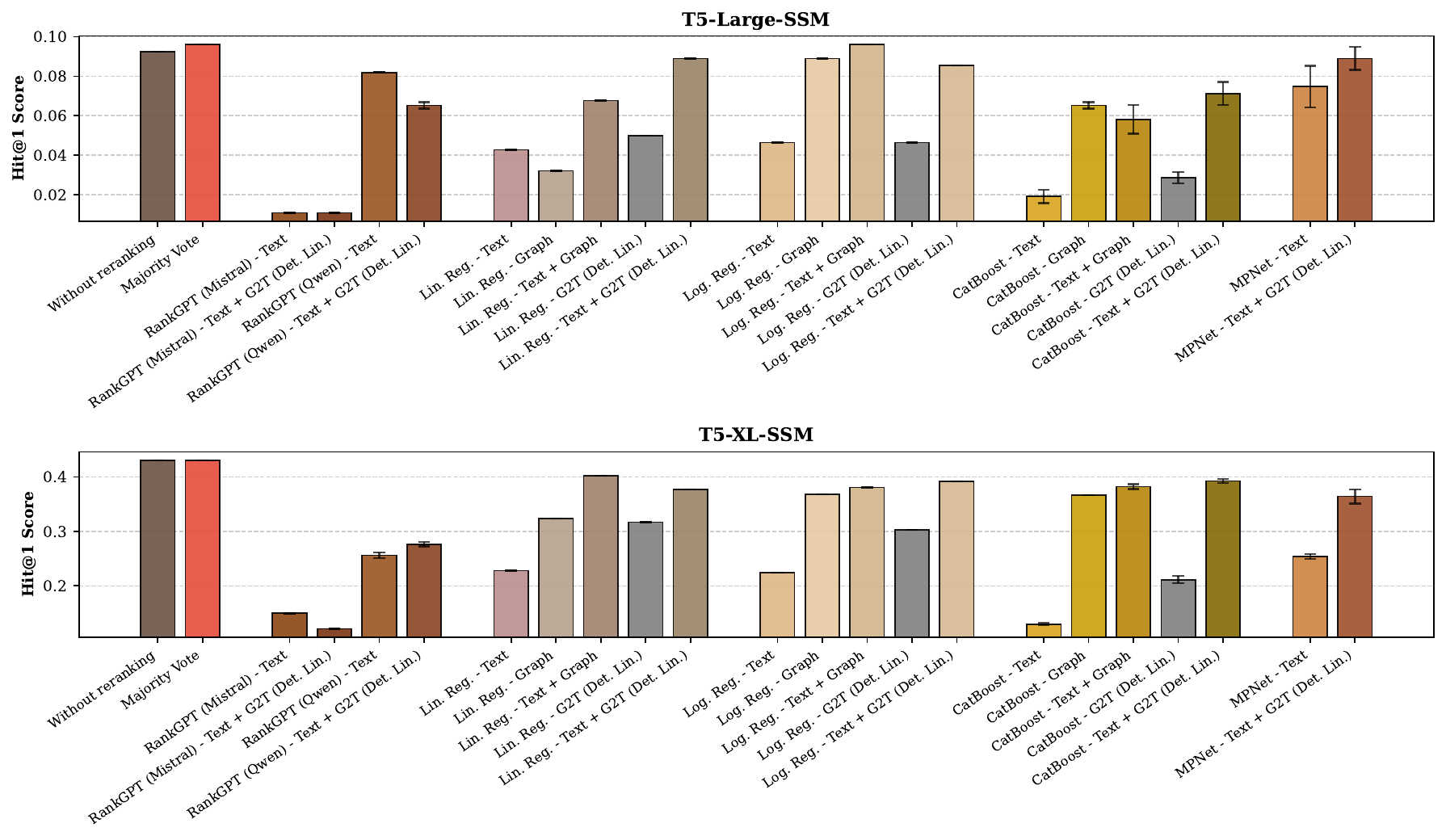}
    \caption{Reranking performance on the MKQA dataset using candidate answers from T5-Large-SSM and T5-XL-SSM. Unlike Mintaka, MKQA requires entity linking for question entities, testing the pipeline's robustness to noisy subgraph extraction.}
    \label{fig:main_results_mkqa_bars}
\end{figure}

This section provides comprehensive results and analysis showcasing the effectiveness of our proposed approach for reranking LLM-generated candidate answers using Knowledge Graph subgraphs. We present results on both the Mintaka and MKQA datasets, evaluating the impact of different combinations of features (graph features and textual representations) on reranking performance and examining the effects of various ranking approaches, ranging from simple to more sophisticated, on reranking Hits@1 accuracy. Detailed tabular results for all experiments, including Hit@2 and Hit@3 metrics, are provided in Appendix~\ref{appx:detailed_results}.

Figure~\ref{fig:main_results_mintaka_bars} and Figure~\ref{fig:main_results_mkqa_bars} present the core experimental results of this work. Figure~\ref{fig:main_results_mintaka_bars} compares the Hit@1 performance on the Mintaka dataset, where gold-standard question entity annotations are available. 
Each subplot corresponds to one of the candidate answer pools and displays the reranking performance of multiple approaches organized by increasing architectural complexity. The approaches include baselines without reranking or using majority voting, RankGPT (a kind of LLM as a Judge), classical regression models (Linear and Logistic Regression), gradient boosting models (CatBoost), and a proposed neural sequence ranker based on MPNet embeddings.
The bars are color-coded by feature type, distinguishing between methods that use only Textual features (question-answer concatenation), Graph features (topological statistics), and Graph2Text (G2T) linearization features, both in deterministic and neural variants. Error bars represent the standard deviation across three independent runs for stochastic methods, providing a measure of model stability.

Figure~\ref{fig:main_results_mkqa_bars} expands this analysis to the MKQA dataset, which presents a more challenging task as it lacks gold entity annotations for questions. This requires our entity linking pipeline to first identify and ground question entities before subgraph extraction. The two subplots in this figure show results for T5-Large-SSM and T5-XL-SSM models, that was used for candidate answer generation. This allows for direct comparison with the Mintaka results, under more realistic conditions where entity linking errors can propagate into the reranking stage. Together, these figures demonstrate the robustness of our approach across different base models, feature representations, and entity linking quality. They also provide clear evidence of the value added by Knowledge Graph-guided reranking.

A critical observation from the figures is the relationship between the base model's quality and the reranking effectiveness. For weaker models such as T5-Large-SSM and T5-XL-SSM, our MPNet-based reranker demonstrates significant improvements over the "Without Reranking" baseline, recovering correct answers that were originally ranked lower in the candidate list. In contrast, for stronger models like Mixtral-8x7B, where the initial Hit@1 score is already high, the benefits from reranking are less significant, and in some cases, simple reranking methods can even slightly reduce performance due to overwriting high-quality initial predictions. This suggests that our KG-guided reranker is particularly useful when the LLM's top-1 predictions are not very reliable.

\subsection{Features Importance}

 \begin{figure}[htb]
  \centering
  \includegraphics[width=0.85\textwidth]{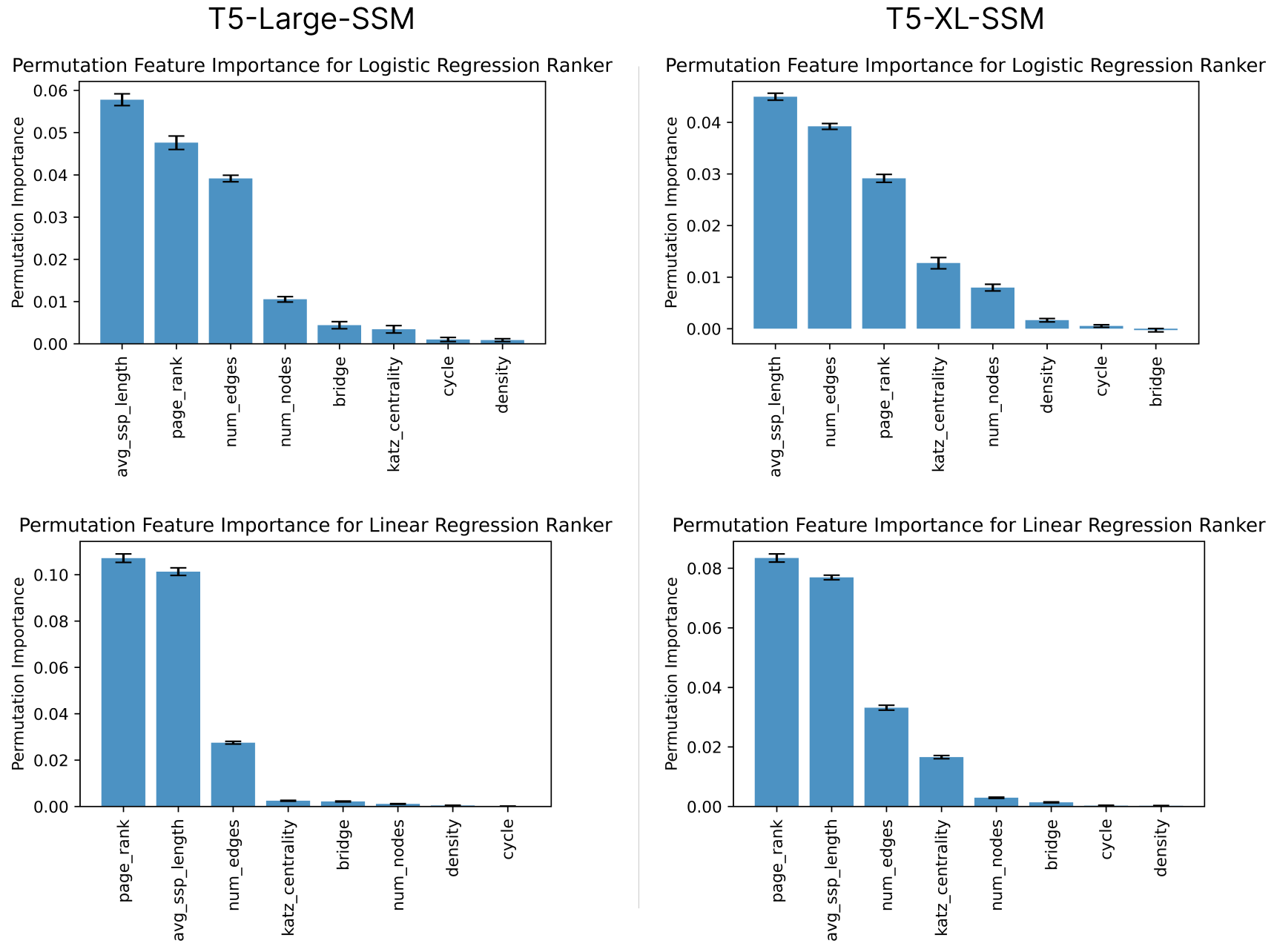}
  \caption{Permutation importance of graph features for Linear and Logistic Regression rankers on answer candidates generated by T5-Large-SSM and by T5-XL-SSM}
  \label{fig:feature_importance_t5_large_xl}
\end{figure}

 \begin{figure}[htb]
  \centering
  \includegraphics[width=0.85\textwidth]{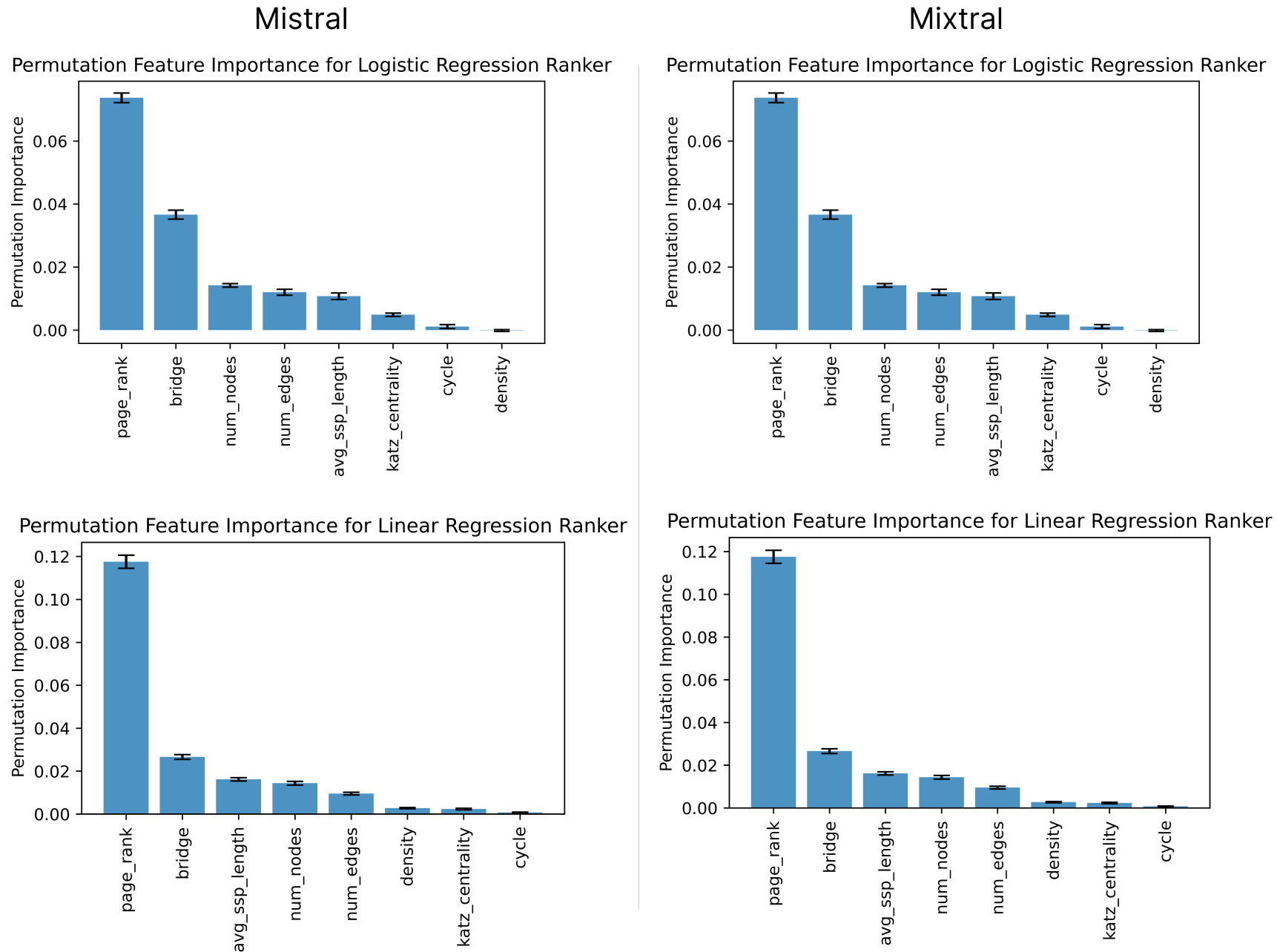}
  \caption{Permutation importance of graph features for Linear and Logistic Regression rankers on answer candidates generated by Mistral and by Mixtral}
  \label{fig:feature_importance_mistral_mixtral}
\end{figure}

In addition to analyzing the overall performance of various reranking approaches, we also investigate the contribution of individual graph features to the ranking decision. To understand which graph-topological properties are most informative in distinguishing between correct and incorrect answers, we calculate permutation feature importance for both regression-based and gradient-boosting rankers.
Note that our best-performing model, the MPNet-based neural network ranker with G2T sequences, uses dense embeddings instead of explicit feature vectors. This makes it difficult to interpret feature-level attributions. Therefore, we focus our analysis on the regression and CatBoost models, where feature importance can be more easily measured.

For regression-based ranking models, we use graph features and embeddings of text and G2T sequences as input. Without the use of embeddings, these models do not achieve competitive performance, suggesting that both structural and semantic information are essential. To enable regression models to handle high-dimensional embeddings, we convert the embedding vectors into separate columns, making it difficult to interpret the importance of each embedding. Therefore, our permutation importance analysis focuses specifically on the more interpretable graph features, which are calculated using 10 repetitions to ensure stability.

Figure~\ref{fig:feature_importance_t5_large_xl} and Figure~\ref{fig:feature_importance_mistral_mixtral} present the permutation feature importance for T5-Large-SSM/T5-XL-SSM and Mistral/Mixtral candidate sources, respectively. The results reveal a consistent pattern across both regression models: PageRank emerges as the dominant feature, significantly outweighing other graph statistics in its contribution to ranking accuracy. This finding validates the intuition that answer candidate entities with higher centrality in the question-answer subgraph are more likely to be correct. Additional features showing notable importance include the number of bridges, node and edge counts, and the average shortest path length between question entities and answer candidates. Interestingly, while the relative ranking of features remains similar across base models, T5-based and Mistral/Mixtral-based candidates exhibit some differences in feature importance magnitudes, suggesting that subgraph characteristics vary depending on the quality and diversity of the candidate pool generated by the underlying LLM.

We extend this feature importance analysis to our gradient-boosting ranker, CatBoost. A key advantage of CatBoost is its native support for embedding features, which allows it to process them as single, unified inputs rather than as flattened vectors of individual values. This enables a direct comparison between the importance of high-level semantic features (Text and Graph2Text embeddings) and the graph features.

\begin{figure}[htb]
  \centering
  \includegraphics[width=\textwidth]{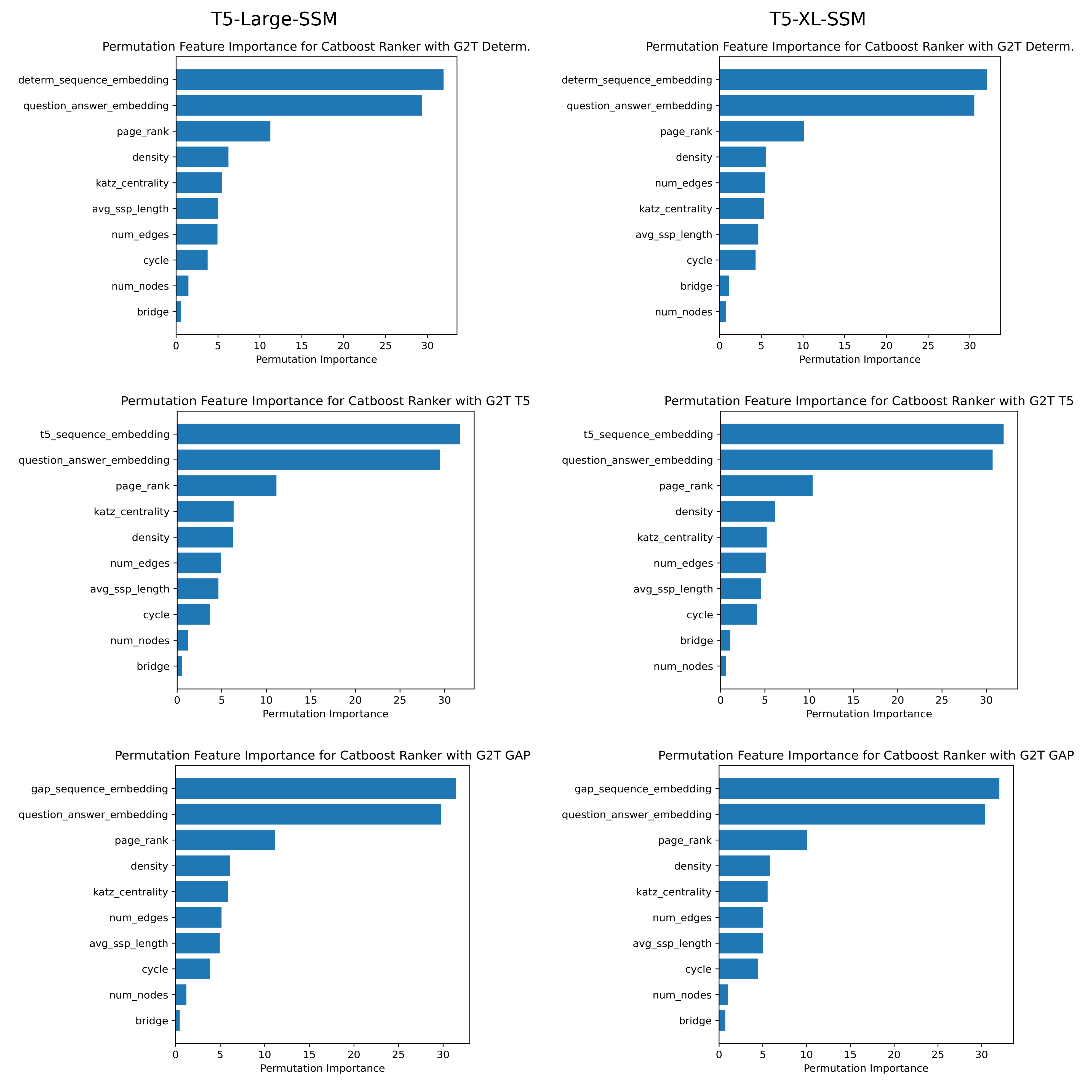}
  \caption{Permutation importance of graph, text, and G2T features for Catboost rankers on answer candidates generated by T5-Large-SSM and T5-XL-SSM}
  \label{fig:feature_importance_t5_large_xl_catboost}
\end{figure}

\begin{figure}[htb]
  \centering
  \includegraphics[width=\textwidth]{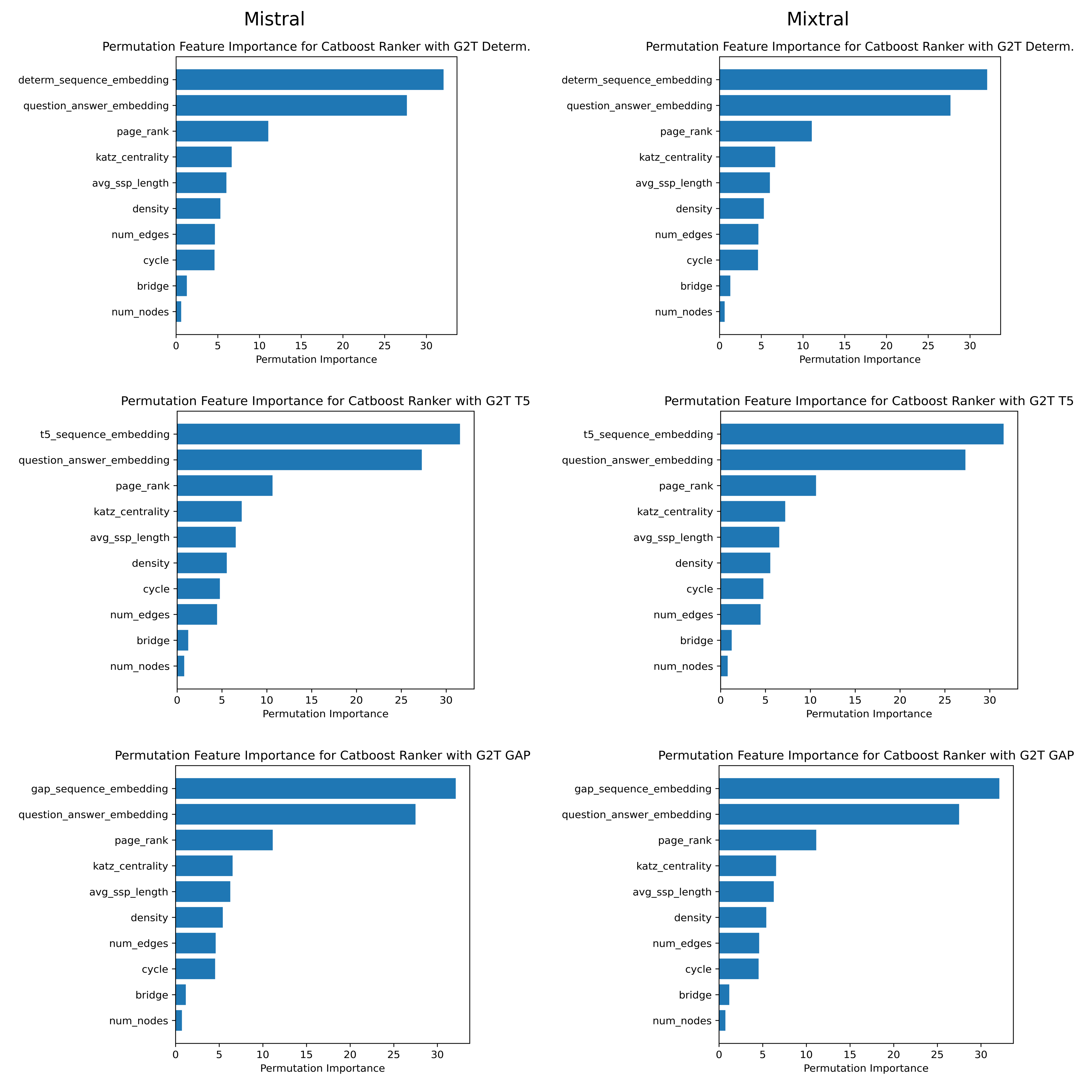}
  \caption{Permutation importance of graph, text, and G2T features for Catboost rankers on answer candidates generated by Mistral and Mixtral}
  \label{fig:feature_importance_mis_mix_catboost}
\end{figure}

To ensure consistency across all experiments, we continue to use embeddings for all textual representations, rather than CatBoost's internal text processing methods. The resulting permutation feature importance scores, comparing the contributions of all graph, text, and G2T features, are presented in Figures~\ref{fig:feature_importance_t5_large_xl_catboost} and ~\ref{fig:feature_importance_mis_mix_catboost}.

Looking more closely at these results, we can see that Graph2Text sequence features and plain text features dominate the importance ranking. This confirms that rich textual representations of the question-candidate pair are the main signal for CatBoost. In particular, including the original question in the textual input has a strong positive effect on the reranking quality.
At the same time, PageRank remains one of the most influential graph features for both T5-based and Mistral/Mixtral-based candidate sets, supporting its usefulness for future KG-based ranking methods. Other structural features such as graph density, Katz centrality, and the average shortest path between question entities and answer candidates also contribute meaningfully. However, the number of bridges plays a much smaller role for CatBoost compared to regression-based rankers.

\subsection{Graph2Text Model Quality Analysis}

Our experiments reveal an unexpected pattern in the performance of neural Graph2Text linearization methods. While both T5-based and GAP-based approaches were pre-trained on the WebNLG 2.0 dataset, where GAP demonstrating superior generation quality based on standard metrics~\cite{DBLP:conf/coling/ColasAW22-GAP}. However, when used as features in downstream reranking, T5-generated sequences consistently yielded better Hit@1 scores. This finding is counterintuitive and motivates a deeper investigation into how these models handle Wikidata subgraphs, which differ structurally from DBpedia graphs in the WebNLG dataset.

\begin{table}[ht]
\caption{Graph2Text Accuracy of label generation for entities in texts generated from subgraphs.}
\label{tab:g2t_accuracy}
\begin{tabular}{llcc}
\toprule
  Answers Source &
   &
  \multicolumn{1}{l}{\begin{tabular}[c]{@{}l@{}}Question Entities\\ Accuracy\end{tabular}} &
  \multicolumn{1}{l}{\begin{tabular}[c]{@{}l@{}}Answer Entities\\ Accuracy\end{tabular}} \\
\midrule
\multicolumn{4}{c}{Graph2Text (T5)}                                   \\
\midrule
\multirow{2}{*}{T5-Large-SSM} & TRAIN & 0.995 & 0.833          \\
                              & TEST  & 0.954 & \textbf{0.835} \\
\hline
\multirow{2}{*}{T5-XL-SSM}    & TRAIN & 0.955 & 0.827          \\
                              & TEST  & 0.955 & 0.829          \\
\midrule
\multicolumn{4}{c}{Graph2Text (GAP)}                                   \\
\midrule
\multirow{2}{*}{T5-Large-SSM} & TRAIN & 0.922 & 0.773          \\
                              & TEST  & 0.923 & 0.776          \\
\hline
\multirow{2}{*}{T5-XL-SSM}    & TRAIN & 0.924 & 0.767          \\
                              & TEST  & 0.926 & 0.77           \\
\hline
\end{tabular}
\end{table}

To understand this, we evaluate the coverage of entities in generated Graph2Text sequences. Specifically, we look at the proportion of entities in the input subgraph that are correctly mentioned in the output text. Our analysis reveals a systematic difference between GAP and T5. GAP tends to omit more entities from the subgraph than T5, which leads to information loss and negatively impacts the ability of the ranker to use the full structural information. Additionally, based on human assessment, GAP has a higher tendency to generate entities or relationships that are not in the input graph, introducing spurious signals that could mislead the ranking model.

These findings suggest that standard Graph2Text evaluation metrics on benchmark datasets may not fully capture the requirements for effective feature generation in knowledge-based reranking tasks. In our study, faithful entity preservation and avoiding hallucination seem to be more critical than the fluency or naturalness of generated text, as the ranking model relies on the presence of specific entity mentions to compute semantic similarity with the question. This observation highlights the importance of task-specific evaluation when selecting Graph2Text models for downstream applications other than pure text generation.

\subsection{Hits@N Evaluation}

\begin{figure}[htbp]
  \centering
  \includegraphics[width=0.99\textwidth]{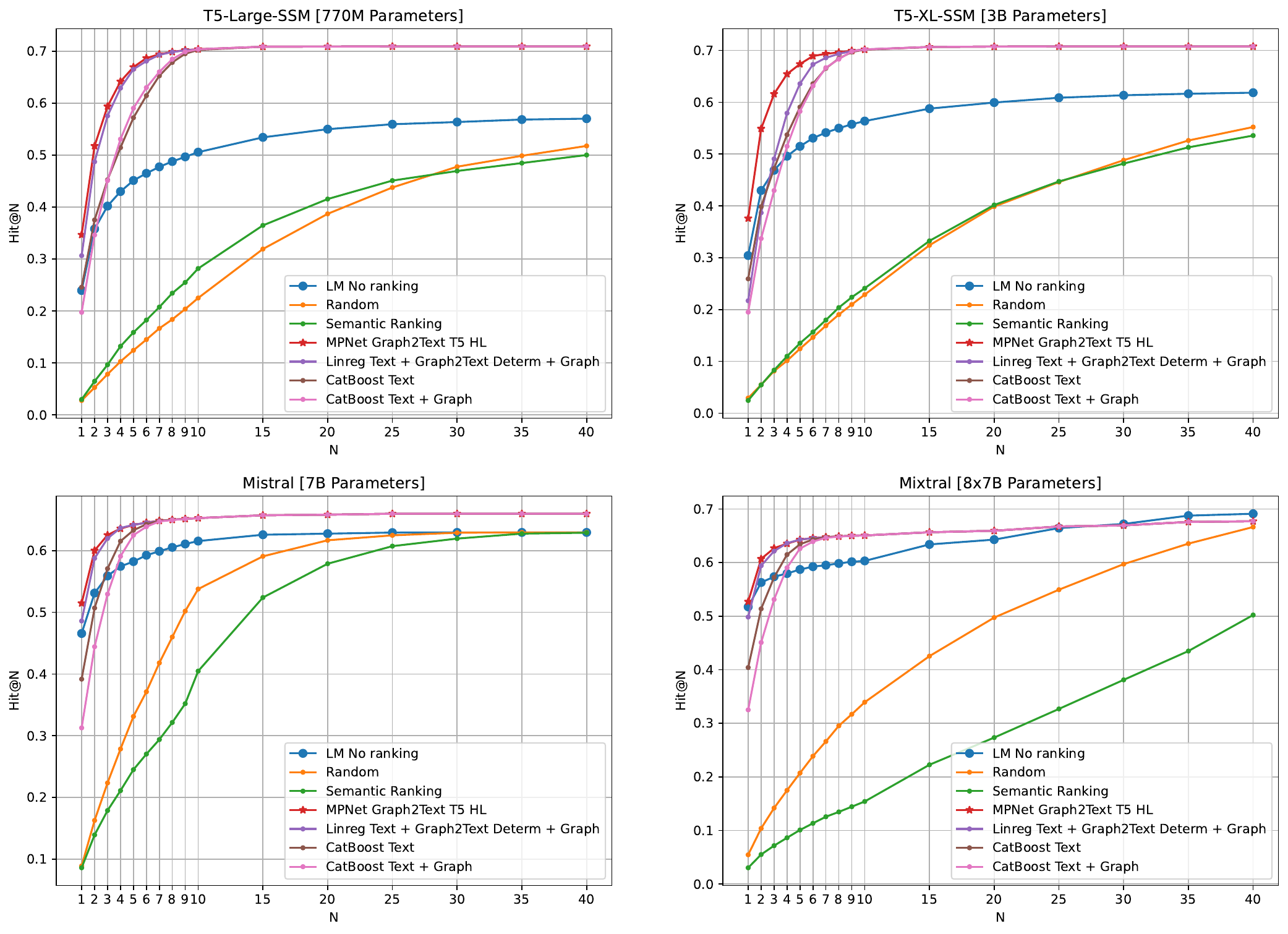}
  \caption{Hit@N Metrics for Different N Values on Mintaka~\cite{DBLP:conf/coling/SenAS22-mintaka} full dataset for generated by Diverse Beam Search~\cite{vijayakumar2018diverse} answers by LLM tuned models with different reranking approaches.}
  \label{fig:hit_at_n_plot_all}
\end{figure}

Solely relying on the Hits@1 or the highest-ranking answer will not give a complete look at the effectiveness of the ranking approaches. As mentioned in \ref{evaluation}, we choose the Hits@N metrics to showcase an exhaustive case study of all rankers and feature sets. That said, we analyze the complete sequence of answers before and after the reranking in this subsection. All proposed methods have been tested on different features, and Hit@N has been calculated for various $N$ values to ensure that the suggested approaches have had an impact. Figure~\ref{fig:hit_at_n_plot_all} shows the most representative and significant results. The reranking produced by MPNet with text and Graph2Text features demonstrates a clear improvement in quality compared to both initial predictions and baseline models. Additionally, we can observe a clear increase in the quality of the Hits@1 answers, regardless of the size of the LLMs.

Furthermore, the results for all possible combinations of answer candidates models, reranking models, feature sets, and answer candidate sources are presented in Appendix \ref{appx:detailed_results}. The results demonstrate the potential of utilizing subgraphs to rerank LLM's responses. Moreover, it is essential to accurately transfer all relevant information from the subgraph to the reranking model. Therefore, the outcomes of this process significantly depend on the quality of the Graph2Text process.

\section{Tool for KGQA and Subgraphs Exploration} \label{sec:webapp}

\begin{figure}[htbp]
  \centering
  \includegraphics[width=0.99\textwidth]{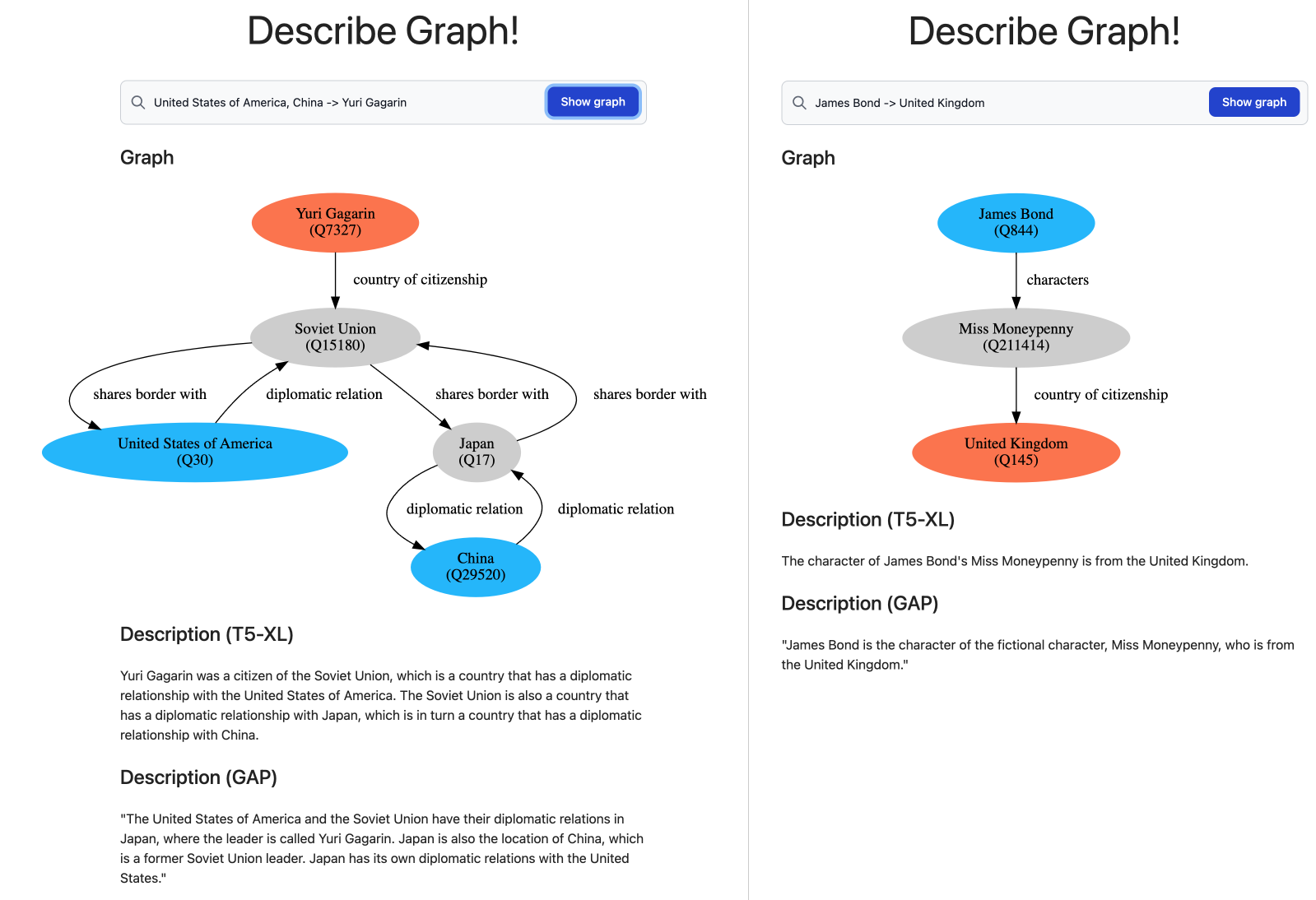}
  \caption{A web application that visualizes subgraphs for given entities. Examples of the USA, China, and Yuri Gagarin are shown on the left side. On the right side are examples of James Bond and the United Kingdom. Both examples provide not only subgraphs but also their Graph2Text representation.}
  \label{fig:webapp_example_viz}
\end{figure}

To explore the space of KGs between arbitrary questions and corresponding answer entities, we have developed a web tool that visualizes subgraphs and automatically applies the Graph2Text (T5 and GAP) methods to these subgraphs. These methods were discussed in Section~\ref{sec:g2t_seq}. In addition, this web application has features for predicting answers to questions and functions as a KGQA system. This web application, available at \url{https://kgqa-nlp-zh.skoltech.ru/graph}, can be used to systematically study subgraphs in detail and evaluate the motivation behind provided answers, as well as explore the space of KGs between questions and answers, which often contains interesting information about knowledge structure.

Figure~\ref{fig:webapp_example_viz} presents two subgraph visualizations that showcase the versatility of our tool in exploring diverse knowledge domains. The Figure is divided into two parts, each demonstrating a unique set of entity relationships.
In the first part of Figure~\ref{fig:webapp_example_viz}, we see a subgraph connecting the countries of the United States, China, and the historical figure Yuri Gagarin. This visualization allows users to explore the complex relationships between countries and important historical figures in the context of space exploration. Interestingly, the shortest path between Yuri Gagarin and China went through Japan. This specific aspect of the KG can be confusing for users and researchers, but techniques like those we provided can help shed light on such matters.
The second part of Figure~\ref{fig:webapp_example_viz} shows a subgraph connecting James Bond and the United Kingdom. This visualization section makes users more interested if a question about James Bond is asked.
By presenting these contrasting examples within a single figure, we demonstrate the tool's capability to visualize and analyze relationships across a wide spectrum of knowledge, from historical events to popular culture, because of general KG - Wikidata~\cite{wikidata}. 

\begin{figure}[htbp]
  \centering
  \includegraphics[width=0.99\textwidth]{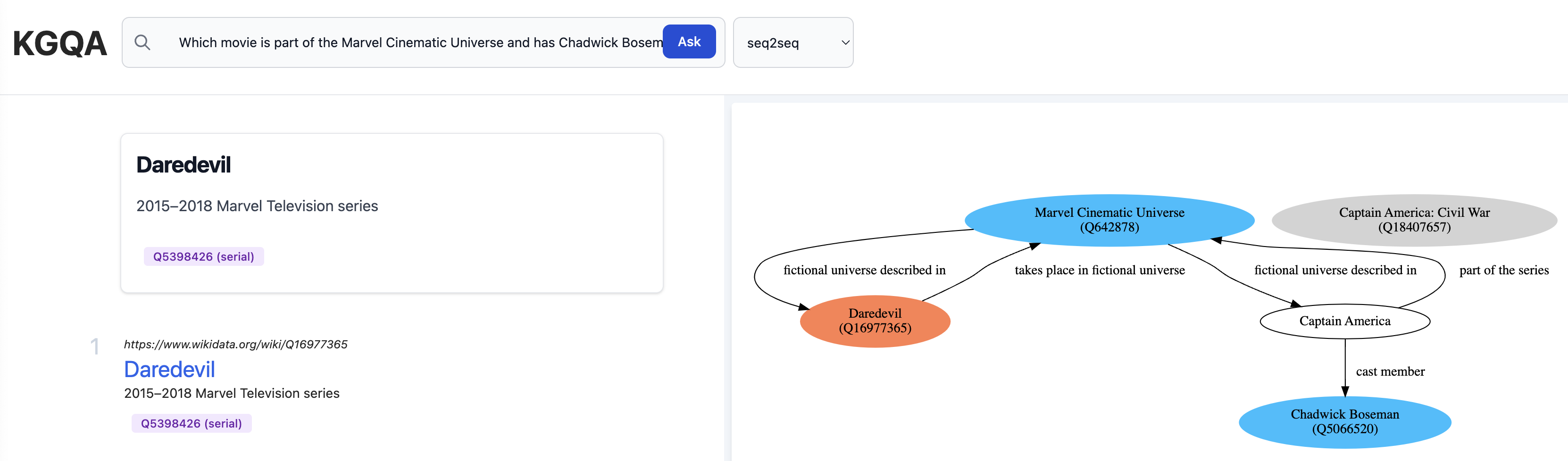}
  \caption{Web application for KGQA with extracted subgraph visualization for question ``\texttt{Which movie is part of the Marvel Cinematic Universe and has Chadwick Boseman in the titular role?}''. Entity linker extracted entity Marvel Cinematic Universe~(Q642878) and Chadwick Boseman~(Q5066520) from question and language model predicted Daredevil~(Q16977365) as a top-1 answer that is incorrect what is indirectly seen from the subgraph.}
  \label{fig:webapp_example_kgqa_bond}
\end{figure}

\begin{figure}[ht]
  \centering
  \includegraphics[width=0.50\textwidth]{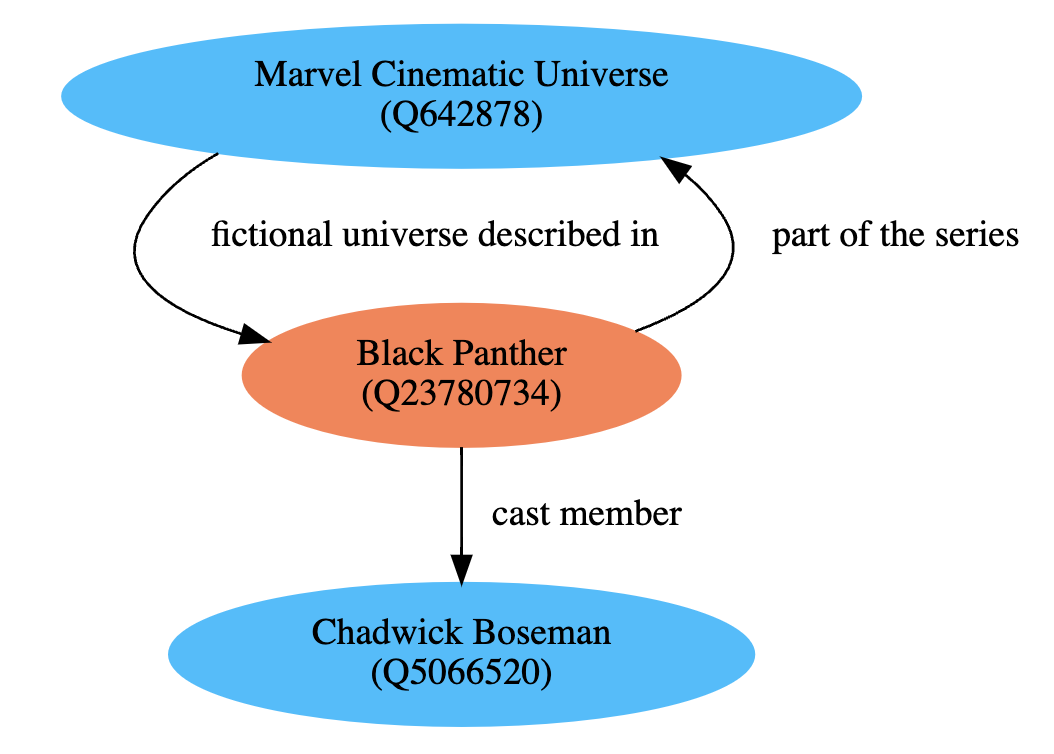}
  \caption{Subgraph for question ``\texttt{Which movie is part of the Marvel Cinematic Universe and has Chadwick Boseman in the titular role?}'' which include entities Marvel Cinematic Universe~(Q642878) and Chadwick Boseman~(Q5066520) in question. The correct answer for the question is entity Black Panther~(Q23780734).}
  \label{fig:webapp_example_kgqa_correct_subgraph}
\end{figure}

We use the FastAPI\footnote{\url{https://fastapi.tiangolo.com}} framework to develop web applications that provide a public API\footnote{\url{https://kgqa-nlp-zh.skoltech.ru/docs}} in addition to web pages. This choice of framework allows for high-performance, easy-to-maintain code with built-in support for asynchronous operations and automatic API documentation generation. To speed up backend processes, we implement a caching mechanism that stores the most popular entities our users interact with. This cache significantly reduces response times for frequently accessed data, enhancing the overall user experience. The caching strategy is dynamically adjusted based on usage patterns and entity popularity, ensuring optimal performance. One of the application's key features is a subgraph visualization in SVG format. Third parties can easily integrate this scalable vector graphics output into their applications or research projects. The web application also provides a KGQA component with various backends, including the T5 sequence-to-sequence model, trained on the Mintaka~\cite{DBLP:conf/coling/SenAS22-mintaka} dataset. 
To illustrate the capabilities of our system, Figure~\ref{fig:webapp_example_kgqa_bond} presents an example query: ``\texttt{Which movie is part of the Marvel Cinematic Universe and has Chadwick Boseman in the titular role?}'' The Figure showcases the answers generated using beam search, a technique that allows exploring multiple potential answer paths. The correct answer for this one question is ``Black Panther'', and the subgraph for this one answer looks simpler and clearer~(Figure~\ref{fig:webapp_example_kgqa_correct_subgraph}) can help the user to be sure about the answer.

\section{Limitations and Future Works}
While our approach has shown improvements in the reranking of LLM-generated answers for KGQA, there are several important limitations that should be acknowledged. In this section, we will discuss the main constraints of our method and propose promising directions for future research.

\subsection{Limitations}
Computational Resource Requirements. Our approach relies on substantial computational resources, which may limit its practicality. Specifically, extracting subgraphs requires loading the entire Wikidata knowledge graph into memory. This process requires approximately 80 GB of RAM per process, making it difficult to deploy in environments with limited resources.

Entity Linking Dependence. The performance of our reranking pipeline is heavily dependent on the accuracy of entity linking. Entity linking is a significant bottleneck in our system, as it can lead to incorrect subgraph extraction and poor reranking results if it fails to identify question entities correctly or produces incorrect Wikidata IDs.

Dataset and Question Type Limitations. Our method is designed to answer factoid questions that have specific entity answers. It does not handle other question types, such as yes/no or count questions, because these questions do not have corresponding answers in Wikidata. This limits the applicability of our method. Additionally, all our experiments were conducted on English-language datasets. The performance of our model on other languages is currently unknown.

\subsection{Future Works}

Extend to Complex Question Types. One area for future research is to address the limitation of handling only entity-based questions. This could be done by developing specialized methods for different types of questions, such as yes/no, count, and temporal reasoning. For yes/no questions, the system could use subgraphs to verify the existence of specific relationships between entities rather than simply ranking candidate entities. For count questions, the graph could be used to estimate the cardinality of the answer and validate the numerical answer generated by the LLM.

Factual Verification for RAG and LLM Systems. Our approach based on subgraphs has great potential for detecting errors and verifying factual accuracy in RAG systems and LLM applications in general. By extracting subgraphs connecting entities mentioned in the generated text by LLMs and analyzing their consistency with facts in the knowledge graph, we can develop automated fact-checking systems. Such systems would be especially valuable for high-stakes applications such as medical diagnosis, legal advice, and scientific research, where accurate facts are critical and errors can have serious consequences.

\section{Conclusion}

In this paper, we propose a novel methodology for reranking answers in KGQA based on the use of subgraphs. We demonstrate the effectiveness of this method by combining it with various feature sets and reranking models and show that incorporating subgraph information can significantly improve accuracy in answer selection. Our work also emphasizes the importance of Semantic Web technologies, which form the basis for Knowledge Graphs. Despite the rise of modern Large Language Models (LLMs), Semantic Web standards remain crucial for creating structured, interpretable, and semantically rich knowledge representations. These technologies enable the creation of Knowledge Graphs, which are essential for tasks such as Question Answering, where precise and reliable information retrieval is essential.

Our work has several implications for further research in the field of KGQA. First, it emphasizes the importance of considering structural relationships between entities to answer complex questions. Second, it demonstrates that integrating graph-based techniques with traditional language models can be beneficial. Finally, our approach suggests that subgraphs may be a valuable tool for enhancing the transparency and interpretability of KGQA systems. 

The proposed methods for reranking answers to factual questions could be implemented in real-world applications, such as zero-click search, a common feature of modern search engines. A web application that visually represents subgraphs of questions and candidate answers, along with a graph-to-text representation, could serve as an additional tool for assisting humans in verifying the accuracy of provided answers.

Our study offers a novel perspective on answer selection in KGQA, emphasizing the potential advantages of incorporating subgraph data into the reranking procedure. We anticipate that this study stimulate further investigation and advancement in this field, ultimately resulting in more precise and efficient question-answering systems.

\newpage

\begin{appendices}

\section{Detailed results for all proposed features and reranking models} \label{appx:detailed_results}

Comprehensive findings for all the suggested feature sets and reranking algorithms are presented.

\begin{table}[htbp]
    \caption{Hit@1-3 for all proposed reranking approaches and features after reranking the T5-Large-SSM generated answer candidates for Mintaka dataset.}
    \label{tab:t5_large_ssm_all_results}
    \fontsize{7pt}{9pt}\selectfont
    \setlength{\tabcolsep}{3pt}
    \begin{tabular}{l p{4cm} c c c}
        \toprule
        \textbf{Reranking Model} & \textbf{Features} & \textbf{Hit@1} & \textbf{Hit@2} & \textbf{Hit@3} \\
        \midrule
        Without reranking & & 0.2395 & 0.3580 & 0.4020 \\
        Full random & & 0.0278 & 0.0562 & 0.0828 \\
        Majority Vote & & 0.2213 & 0.3422 & 0.3875 \\
        Semantic reranking & Text & 0.0298 & 0.0647 & 0.0965 \\
        \cmidrule(lr){1-5}
        \multirow{2}{*}{RankGPT (Mistral)} & Text & $0.2442\pm0.0004$ & $0.3237\pm0.0004$ & $0.3864\pm0.0007$ \\
        & Text + G2T (Det. Lin.) & $0.1984\pm0.0007$ & $0.2904\pm0.0008$ & $0.3613\pm0.0010$ \\
        \cmidrule(lr){1-5}
        \multirow{2}{*}{RankGPT (Qwen2.5 7B)} & Text & $0.4141\pm0.0003$ & $0.5053\pm0.0007$ & $0.5445\pm0.0000$ \\
         & Text + G2T (Det. Lin.) & $0.3933\pm0.0003$ & $0.4952\pm0.0003$ & $0.5388\pm0.0002$ \\
        \cmidrule(lr){1-5}
        \multirow{9}{*}{Linear Regression} & Text & 0.2695 & 0.4502 & 0.5400 \\
        & Graph & 0.2338 & 0.4007 & 0.5000 \\
        & Text + Graph & 0.2953 & 0.4697 & 0.5558 \\
        & G2T (Det. Lin.) & 0.2550 & 0.4262 & 0.5230 \\
        & G2T (T5) & 0.2505 & 0.4230 & 0.5193 \\
        & G2T (GAP) & 0.2393 & 0.4133 & 0.5123 \\
        & Text + Graph + G2T (Det. Lin.) & \textbf{0.3065} & 0.4873 & 0.5755 \\
        & Text + Graph + G2T (T5) & 0.3070 & 0.4835 & 0.5710 \\
        & Text + Graph + G2T (GAP) & 0.3033 & 0.4870 & 0.5728 \\
        \midrule
        \multirow{12}{*}{Logistic Regression} & Text & 0.2605 & 0.4353 & 0.5360 \\
        & Graph & 0.2335 & 0.4040 & 0.4980 \\
        & Text + Graph & 0.2470 & 0.4240 & 0.5218 \\
        & G2T (Det. Lin.) & 0.2440 & 0.4203 & 0.5163 \\
        & G2T (T5) & 0.2313 & 0.4068 & 0.5028 \\
        & G2T (GAP) & 0.2925 & 0.4688 & 0.5625 \\
        & Text + Graph + G2T (Det. Lin.) & \textbf{0.3060} & 0.4873 & 0.5745 \\
        & Text + Graph + G2T (T5) & 0.2794 & 0.4780 & 0.5668 \\
        & Text + Graph + G2T (GAP) & 0.2925 & 0.4688 & 0.5625 \\
        \midrule
        \multirow{12}{*}{CatBoost} & Text & $0.2421\pm0.0034$ & $0.3784\pm0.0068$ & $0.4512\pm0.0061$ \\
        & Graph & $0.2614\pm0.0008$ & $0.4292\pm0.0011$ & $0.5213\pm0.0011$ \\
        & Text + Graph & $0.3125\pm0.0000$ & $0.4632\pm0.0000$ & $0.5370\pm0.0000$ \\
        & G2T (Det. Lin.) & $0.2437\pm0.0008$ & $0.4072\pm0.0011$ & $0.5002\pm0.0011$ \\
        & G2T (T5) & $0.2357\pm0.0008$ & $0.4068\pm0.0011$ & $0.4940\pm0.0011$ \\
        & G2T (GAP) & $0.2357\pm0.0008$ & $0.4068\pm0.0011$ & $0.4940\pm0.0011$ \\
        & Text + Graph + G2T (Det. Lin.) & $0.3144\pm0.0008$ & $0.4765\pm0.0011$ & $0.5575\pm0.0011$ \\
        & Text + Graph + G2T (T5) & $0.3074\pm0.0008$ & $0.4780\pm0.0011$ & $0.5540\pm0.0011$ \\
        & Text + Graph + G2T (GAP) & $0.3091\pm0.0008$ & $0.4658\pm0.0011$ & $0.5535\pm0.0011$ \\
        \midrule
        \multirow{10}{*}{MPNet} & Text & $0.3663\pm0.0027$ & $0.5438\pm0.0015$ & $0.6128\pm0.0018$ \\
        & Text + G2T (Det. Lin.) & $0.3897\pm0.0007$ & $0.5670\pm0.0012$ & $0.6270\pm0.0008$ \\
        & Text + G2T (Det. Lin.) + HL & $0.3867\pm0.0022$ & $0.5666\pm0.0014$ & $0.6322\pm0.0014$ \\
        & Text + G2T (T5) & $0.3816\pm0.0031$ & $0.5583\pm0.0026$ & $0.6243\pm0.0018$ \\
        & Text + G2T (T5) + HL & $0.2607\pm0.1844$ & $0.3749\pm0.2651$ & $0.4189\pm0.2962$ \\
        & Text + G2T (GAP) & $0.3737\pm0.0035$ & $0.5517\pm0.0051$ & $0.6190\pm0.0025$ \\
        & Text + G2T (GAP) + HL & $0.2519\pm0.1781$ & $0.3715\pm0.2627$ & $0.4157\pm0.2939$ \\
        \bottomrule
    \end{tabular}
\end{table}
\begin{table}[htbp]
    \setlength{\tabcolsep}{3pt}
    \caption{Hit@1-3 for all proposed reranking approaches and features after reranking the T5-XL-SSM generated answer candidates for Mintaka Dataset.}
    \label{tab:t5_xl_ssm_all_results}
    \fontsize{7pt}{9pt}\selectfont
    \begin{tabular}{l p{4cm} c c c}
        \toprule
        \textbf{Reranking Model} & \textbf{Features} & \textbf{Hit@1} & \textbf{Hit@2} & \textbf{Hit@3} \\
        Without reranking & & 0.3042 & 0.4298 & 0.4688 \\
        Full random & & 0.0235 & 0.0473 & 0.0723 \\
        Majority Vote & & 0.2677 & 0.4020 & 0.4417 \\
        Semantic reranking & Text & 0.0245 & 0.0545 & 0.0830 \\
        \cmidrule(lr){1-5}
        \multirow{2}{*}{RankGPT (Mistral)} & Text & $0.2501\pm0.0011$ & $0.3356\pm0.0001$ & $0.4017\pm0.0007$ \\
         & Text + G2T (Det. Lin.) & $0.2083\pm0.0004$ & $0.3058\pm0.0004$ & $0.3791\pm0.0007$ \\
        \cmidrule(lr){1-5}
        \multirow{2}{*}{RankGPT (Qwen2.5 7B)} & Text & $0.3877\pm0.0006$ & $0.4909\pm0.0012$ & $0.5370\pm0.0007$ \\
         & Text + G2T (Det. Lin.) & $0.3777\pm0.0004$ & $0.4883\pm0.0007$ & $0.5383\pm0.0002$ \\
        \cmidrule(lr){1-5}
        \multirow{9}{*}{Linear Regression} & Text & 0.2955 & 0.4763 & 0.4763 \\
        & Graph & 0.2550 & 0.4353 & 0.5285 \\
        & Text + Graph & 0.3003 & 0.4825 & 0.5693 \\
        & G2T (Det. Lin.) & 0.2640 & 0.4480 & 0.5375 \\
        & G2T (T5) & 0.2593 & 0.4438 & 0.5333 \\
        & G2T (GAP) & 0.2563 & 0.4320 & 0.5215 \\
        & Text + Graph + G2T (Det. Lin.) & \textbf{0.3220} & 0.5073 & 0.5915 \\
        & Text + Graph + G2T (T5) & 0.3170 & 0.5035 & 0.5818 \\
        & Text + Graph + G2T (GAP) & 0.3153 & 0.5023 & 0.5793 \\
        \midrule
        \multirow{12}{*}{Logistic Regression} & Text & 0.2850 & 0.4683 & 0.5603 \\
        & Graph & 0.2613 & 0.4385 & 0.5283 \\
        & Text + Graph & 0.3003 & 0.4825 & 0.5693 \\
        & G2T (Det. Lin.) & 0.2598 & 0.4420 & 0.5423 \\
        & G2T (T5) & 0.2538 & 0.4393 & 0.5315 \\
        & G2T (GAP) & 0.2503 & 0.4310 & 0.5218 \\
        & Text + Graph + G2T (Det. Lin.) & \textbf{0.3210} & 0.5025 & 0.5885 \\
        & Text + Graph + G2T (T5) & 0.3080 & 0.4955 & 0.5778 \\
        & Text + Graph + G2T (GAP) & 0.3043 & 0.4908 & 0.5815 \\
        \midrule
        \multirow{12}{*}{CatBoost} & Text & $0.2821\pm0.0030$ & $0.4351\pm0.0024$ & $0.5121\pm0.0018$ \\
        & Graph & $0.2786\pm0.0015$ & $0.4575\pm0.0007$ & $0.5475\pm0.0004$ \\
        & Text + Graph & $0.3232\pm0.0055$ & $0.4877\pm0.0038$ & $0.5708\pm0.0065$ \\
        & G2T (Det. Lin.) & $0.2348\pm0.0139$ & $0.4050\pm0.0145$ & $0.4951\pm0.0072$ \\
        & G2T (T5) & $0.2347\pm0.0105$ & $0.4013\pm0.0145$ & $0.4928\pm0.0075$ \\
        & G2T (GAP) & $0.2347\pm0.0055$ & $0.3945\pm0.0085$ & $0.4865\pm0.0060$ \\
        & Text + Graph + G2T (Det. Lin.) & $0.3266\pm0.0090$ & $0.4961\pm0.0104$ & $0.5669\pm0.0115$ \\
        & Text + Graph + G2T (T5) & $0.3271\pm0.0086$ & $0.4907\pm0.0115$ & $0.5692\pm0.0123$ \\
        & Text + Graph + G2T (GAP) & $0.3237\pm0.0108$ & $0.4957\pm0.0094$ & $0.5686\pm0.0107$ \\
        \midrule
        \multirow{10}{*}{MPNet} & Text & $0.3705\pm0.0017$ & $0.5487\pm0.0015$ & $0.6165\pm0.0011$ \\
        & Text + G2T (Det. Lin.) & $0.3907\pm0.0045$ & $0.5673\pm0.0044$ & $0.6278\pm0.0031$ \\
        & Text + G2T (Det. Lin.) + HL & $0.3950\pm0.0036$ & $0.5719\pm0.0038$ & $0.6286\pm0.0015$ \\
        & Text + G2T (T5) & $0.3834\pm0.0011$ & $0.5586\pm0.0007$ & $0.6234\pm0.0014$ \\
        & Text + G2T (T5) + HL & $0.3886\pm0.0024$ & $0.5653\pm0.0012$ & $0.6262\pm0.0007$ \\
        & Text + G2T (GAP) & $0.3808\pm0.0026$ & $0.5551\pm0.0015$ & $0.6159\pm0.0008$ \\
        & Text + G2T (GAP) + HL & $0.3897\pm0.0047$ & $0.5653\pm0.0043$ & $0.6228\pm0.0007$ \\
        \bottomrule
    \end{tabular}
\end{table}

\begin{table}[htbp]
    \setlength{\tabcolsep}{3pt}
    \caption{Hit@1-3 for all proposed reranking approaches and features after reranking the Mistral generated answer candidates for Mintaka dataset.}
    \label{tab:mistral_all_results}
    \fontsize{7pt}{9pt}\selectfont
    \begin{tabular}{l p{4cm} c c c}
        \toprule
        \textbf{Reranking Model} & \textbf{Features} & \textbf{Hit@1} & \textbf{Hit@2} & \textbf{Hit@3} \\
        \midrule
        Without reranking & & 0.4655 & 0.5313 & 0.5590 \\
        Full random & & 0.0920 & 0.1585 & 0.2165 \\
        Majority Vote & & 0.4228 & 0.5020 & 0.5345 \\
        Semantic reranking & Text & 0.0858 & 0.1393 & 0.1790 \\
        \cmidrule(lr){1-5}
        \multirow{2}{*}{RankGPT (Mistral)} & Text & $0.3838\pm0.0007$ & $0.4618\pm0.0002$ & $0.5112\pm0.0004$ \\
         & Text + G2T (Det.Lin.) & $0.3315\pm0.0006$ & $0.4310\pm0.0005$ & $0.4893\pm0.0005$ \\
        \cmidrule(lr){1-5}
        \multirow{2}{*}{RankGPT (Qwen2.5 7B)} & Text & $0.5097\pm0.0003$ & $0.5762\pm0.0002$ & $0.5943\pm0.0002$ \\
         & Text + G2T (Det.Lin.) & $0.5042\pm0.0003$ & $0.5717\pm0.0000$ & $0.5914\pm0.0001$ \\
        \cmidrule(lr){1-5}
        \multirow{9}{*}{Linear Regression} & Text & 0.4760 & 0.5818 & 0.6173 \\
        & Graph & 0.3575 & 0.4918 & 0.5653 \\
        & Text + Graph & 0.4810 & 0.5848 & 0.6193 \\
        & G2T (Det. Lin.) & 0.3960 & 0.5298 & 0.5898 \\
        & G2T (T5) & 0.4013 & 0.5295 & 0.5910 \\
        & G2T (GAP) & 0.3885 & 0.5220 & 0.5845 \\
        & Text + G2T (Det. Lin.) & 0.4860 & 0.5875 & 0.6198 \\
        & Text + G2T (T5) & 0.4825 & 0.5845 & 0.6200 \\
        & Text + G2T (GAP) & 0.4770 & 0.5798 & 0.6180 \\
        \midrule
        \multirow{12}{*}{Logistic Regression} & Text & 0.4730 & 0.5795 & 0.6148 \\
        & Graph & 0.3558 & 0.4925 & 0.5645 \\
        & Text + Graph & 0.4745 & 0.5833 & 0.6200 \\
        & G2T (Det. Lin.) & 0.3970 & 0.5260 & 0.5960 \\
        & G2T (T5) & 0.3985 & 0.5305 & 0.5933 \\
        & G2T (GAP) & 0.3850 & 0.5195 & 0.5878 \\
        & Text + G2T (Det. Lin.) & 0.4790 & 0.5833 & 0.6210 \\
        & Text + G2T (T5) & 0.4745 & 0.5838 & 0.6205 \\
        & Text + G2T (GAP) & 0.4728 & 0.5785 & 0.6183 \\
        \midrule
        \multirow{12}{*}{CatBoost} & Text & $0.3905\pm0.0015$ & $0.5033\pm0.0031$ & $0.5709\pm0.0015$ \\
        & Graph & $0.3705\pm0.0012$ & $0.4990\pm0.0008$ & $0.5706\pm0.0004$ \\
        & Text + Graph & $0.4258\pm0.0007$ & $0.5417\pm0.0002$ & $0.5952\pm0.0009$ \\
        & G2T (Det. Lin.) & $0.3913\pm0.0034$ & $0.5156\pm0.0012$ & $0.5793\pm0.0001$ \\
        & G2T (T5) & $0.3755\pm0.0049$ & $0.5079\pm0.0016$ & $0.5775\pm0.0002$ \\
        & G2T (GAP) & $0.3791\pm0.0022$ & $0.5064\pm0.0016$ & $0.5743\pm0.0028$ \\
        & Text + G2T (Det. Lin.) & $0.4370\pm0.0025$ & $0.5475\pm0.0025$ & $0.5988\pm0.0008$ \\
        & Text + G2T (T5) & $0.4333\pm0.0016$ & $0.5460\pm0.0014$ & $0.5970\pm0.0011$ \\
        & Text + G2T (GAP) & $0.4342\pm0.0011$ & $0.5483\pm0.0025$ & $0.5995\pm0.0030$ \\
        \midrule
        \multirow{10}{*}{MPNet} & Text & $0.5087\pm0.0033$ & $0.5985\pm0.0017$ & $0.6276\pm0.0008$ \\
        & Text + G2T (Det. Lin.) & $0.4956\pm0.0013$ & $0.5931\pm0.0024$ & $0.6245\pm0.0013$ \\
        & Text + G2T (Det. Lin.) + HL & $0.5146\pm0.0045$ & $0.5992\pm0.0021$ & $0.6267\pm0.0006$ \\
        & Text + G2T (T5) & $0.4978\pm0.0022$ & $0.5928\pm0.0012$ & $0.6251\pm0.0017$ \\
        & Text + G2T (T5) + HL & $0.5100\pm0.0018$ & $0.6018\pm0.0020$ & $0.6282\pm0.0013$ \\
        & Text + G2T (GAP) & $0.4955\pm0.0013$ & $0.5911\pm0.0015$ & $0.6232\pm0.0007$ \\
        & Text + G2T (GAP) + HL & $0.5104\pm0.0006$ & $0.6002\pm0.0010$ & $0.6279\pm0.0001$ \\
        \bottomrule
    \end{tabular}
\end{table}

\begin{table}[htbp]
    \setlength{\tabcolsep}{3pt}
    \caption{Hit@1-3 for all proposed reranking approaches and features after reranking the Mixtral generated answer candidates for Mintaka Dataset.}
    \label{tab:mixtral_all_results}
    \fontsize{7pt}{9pt}\selectfont
    \begin{tabular}{l p{4cm} c c c}
        \toprule
        \textbf{Reranking Model} & \textbf{Features} & \textbf{Hit@1} & \textbf{Hit@2} & \textbf{Hit@3} \\
        \midrule
        Without reranking & & 0.5173 & 0.5628 & 0.5735 \\
        Full random & & 0.0580 & 0.1090 & 0.1455 \\
        Majority Vote & & 0.4858 & 0.5645 & 0.5930 \\
        Semantic reranking & Text & 0.0303 & 0.0550 & 0.0715 \\
        \cmidrule(lr){1-5}
        \multirow{2}{*}{RankGPT (Mistral)} & Text & $0.3958\pm0.0005$ & $0.4733\pm0.0002$ & $0.5233\pm0.0004$ \\
         & Text + G2T (Det. Lin.) & $0.3430\pm0.0007$ & $0.4422\pm0.0001$ & $0.5012\pm0.0003$ \\
        \cmidrule(lr){1-5}
        \multirow{2}{*}{RankGPT (Qwen2.5 7B)} & Text & $0.5217\pm0.0003$ & $0.5880\pm0.0002$ & $0.6065\pm0.0002$ \\
         & Text + G2T (Det.Lin.) & $0.5162\pm0.0004$ & $0.5839\pm0.0001$ & $0.6038\pm0.0001$ \\
        \cmidrule(lr){1-5}
        \multirow{9}{*}{Linear Regression} & Text & 0.4883 & 0.5883 & 0.6188 \\
        & Graph & 0.3698 & 0.4983 & 0.5668 \\
        & Text + Graph & 0.4933 & 0.5913 & 0.6208 \\
        & G2T (Det. Lin.) & 0.4083 & 0.5363 & 0.5913 \\
        & G2T (T5) & 0.4135 & 0.5360 & 0.5925 \\
        & G2T (GAP) & 0.4008 & 0.5285 & 0.5860 \\
        & Text + G2T (Det. Lin.) & 0.4983 & 0.5940 & 0.6213 \\
        & Text + G2T (T5) & 0.4948 & 0.5910 & 0.6215 \\
        & Text + G2T (GAP) & 0.4850 & 0.5850 & 0.6198 \\
        \cmidrule(lr){1-5}
        \multirow{12}{*}{Logistic Regression} & Text & 0.4853 & 0.5860 & 0.6163 \\
        & Graph & 0.3680 & 0.4990 & 0.5660 \\
        & Text + Graph & 0.4868 & 0.5898 & 0.6215 \\
        & G2T (Det. Lin.) & 0.4093 & 0.5325 & 0.5975 \\
        & G2T (T5) & 0.4108 & 0.5370 & 0.5948 \\
        & G2T (GAP) & 0.3973 & 0.5260 & 0.5893 \\
        & Text + G2T (Det. Lin.) & 0.4913 & 0.5898 & 0.6225 \\
        & Text + G2T (T5) & 0.4868 & 0.5903 & 0.6220 \\
        & Text + G2T (GAP) & 0.4850 & 0.5850 & 0.6198 \\
        \cmidrule(lr){1-5}
        \multirow{12}{*}{CatBoost} & Text & $0.4020\pm0.0015$ & $0.5093\pm0.0031$ & $0.5722\pm0.0015$ \\
        & Graph & $0.3820\pm0.0012$ & $0.5050\pm0.0008$ & $0.5718\pm0.0004$ \\
        & Text + Graph & $0.4376\pm0.0005$ & $0.5480\pm0.0000$ & $0.5968\pm0.0007$ \\
        & G2T (Det. Lin.) & $0.4033\pm0.0034$ & $0.5221\pm0.0012$ & $0.5811\pm0.0001$ \\
        & G2T (T5) & $0.3870\pm0.0049$ & $0.5139\pm0.0016$ & $0.5787\pm0.0002$ \\
        & G2T (GAP) & $0.3906\pm0.0022$ & $0.5124\pm0.0016$ & $0.5755\pm0.0028$ \\
        & Text + G2T (Det. Lin.) & $0.4485\pm0.0025$ & $0.5535\pm0.0025$ & $0.6001\pm0.0008$ \\
        & Text + G2T (T5) & $0.4448\pm0.0016$ & $0.5520\pm0.0014$ & $0.5983\pm0.0011$ \\
        & Text + G2T (GAP) & $0.4457\pm0.0011$ & $0.5543\pm0.0025$ & $0.6008\pm0.0030$ \\
        \cmidrule(lr){1-5}
        \multirow{10}{*}{MPNet} & Text & $0.5162\pm0.0002$ & $0.6042\pm0.0015$ & $0.6286\pm0.0006$ \\
        & Text + G2T (Det. Lin.) & $0.5074\pm0.0007$ & $0.5989\pm0.0013$ & $0.6248\pm0.0020$ \\
        & Text + G2T (Det. Lin.) + HL & $0.5268\pm0.0037$ & $0.6049\pm0.0022$ & $0.6279\pm0.0016$ \\
        & Text + G2T (T5) & $0.5067\pm0.0012$ & $0.5965\pm0.0009$ & $0.6248\pm0.0012$ \\
        & Text + G2T (T5) + HL & $0.5199\pm0.0005$ & $0.6074\pm0.0014$ & $0.6301\pm0.0007$ \\
        & Text + G2T (GAP) & $0.5057\pm0.0012$ & $0.5944\pm0.0015$ & $0.6231\pm0.0008$ \\
        & Text + G2T (GAP) + HL & $0.5225\pm0.0015$ & $0.6058\pm0.0010$ & $0.6286\pm0.0010$ \\
        \bottomrule
    \end{tabular}
\end{table}

\begin{table}[htbp]
    \setlength{\tabcolsep}{3pt}
    \caption{Hit@1-3 for all proposed reranking approaches and features after reranking the T5 Large SSM generated answer candidates on MKQA dataset.}
    \label{tab:mkqa_t5_large_ssm_all_results}
    \fontsize{7pt}{9pt}\selectfont
    \begin{tabular}{l p{4cm} c c c}
        \toprule
        \textbf{Reranking Model} & \textbf{Features} & \textbf{Hit@1} & \textbf{Hit@2} & \textbf{Hit@3} \\
        \midrule
        Without reranking & & 0.0925 & 0.1139 & 0.1317 \\
        Full random & & 0.0249 & 0.0391 & 0.0712 \\
        Majority Vote & & 0.0961 & 0.1281 & 0.1423 \\
        Semantic reranking & Text & 0.0000 & 0.0000 & 0.0000 \\
        \cmidrule(lr){1-5}
        \multirow{2}{*}{RankGPT (Mistral)} & Text & $0.0107\pm0.0000$ & $0.0463\pm0.0000$ & $0.0747\pm0.0000$ \\
        & Text + G2T (Det. Lin.) & $0.0107\pm0.0000$ & $0.0356\pm0.0000$ & $0.0569\pm0.0000$ \\
        \cmidrule(lr){1-5}
        \multirow{2}{*}{RankGPT (Qwen2.5 7B)} & Text & $0.0819\pm0.0000$ & $0.1246\pm0.0029$ & $0.1447\pm0.0017$ \\
        & Text + G2T (Det. Lin.) & $0.0652\pm0.0017$ & $0.1056\pm0.0017$ & $0.1340\pm0.0017$ \\
        \cmidrule(lr){1-5}
        \multirow{5}{*}{Linear Regression} & Text & 0.0427 & 0.0641 & 0.0996 \\
        & Graph & 0.0320 & 0.0534 & 0.0854 \\
        & Text + Graph & 0.0676 & 0.0961 & 0.1246 \\
        & G2T (Det. Lin.) & 0.0498 & 0.0747 & 0.1068 \\
        & Text + G2T (Det. Lin.) & 0.0890 & 0.1459 & 0.1673 \\
        \midrule
        \multirow{5}{*}{Logistic Regression} & Text & 0.0463 & 0.0747 & 0.0996 \\
        & Graph & 0.0890 & 0.1352 & 0.1530 \\
        & Text + Graph & 0.0961 & 0.1317 & 0.1708 \\
        & G2T (Det. Lin.) & 0.0463 & 0.0712 & 0.0925 \\
        & Text + G2T (Det. Lin.) & 0.0854 & 0.1139 & 0.1566 \\
        \midrule
        \multirow{5}{*}{CatBoost} & Text & $0.0190\pm0.0034$ & $0.0261\pm0.0017$ & $0.0498\pm0.0058$ \\
        & Graph & $0.0652\pm0.0017$ & $0.1020\pm0.0044$ & $0.1281\pm0.0000$ \\
        & Text + Graph & $0.0581\pm0.0073$ & $0.0985\pm0.0093$ & $0.1234\pm0.0073$ \\
        & G2T (Det. Lin.) & $0.0285\pm0.0029$ & $0.0581\pm0.0060$ & $0.0688\pm0.0017$ \\
        & Text + G2T (Det. Lin.) & $0.0712\pm0.0058$ & $0.0985\pm0.0117$ & $0.1222\pm0.0017$ \\
        \midrule
        \multirow{2}{*}{MPNet} & Text & $0.0747\pm0.0105$ & $0.1056\pm0.0143$ & $0.1317\pm0.0151$ \\
        & Text + G2T (Det. Lin.) & $0.0890\pm0.0058$ & $0.1340\pm0.0170$ & $0.1590\pm0.0143$ \\
        \bottomrule
    \end{tabular}
\end{table}

\begin{table}[htbp]
    \setlength{\tabcolsep}{3pt}
    \caption{Hit@1-3 for all proposed reranking approaches and features after reranking the T5 XL SSM generated answer candidates on MKQA Dataset.}
    \label{tab:mkqa_t5_xl_ssm_all_results}
    \fontsize{7pt}{9pt}\selectfont
    \begin{tabular}{l p{4cm} c c c}
        \toprule
        \textbf{Reranking Model} & \textbf{Features} & \textbf{Hit@1} & \textbf{Hit@2} & \textbf{Hit@3} \\
        \midrule
        Without reranking & & 0.4306 & 0.4911 & 0.5267 \\
        Full random & & 0.2112 & 0.2871 & 0.3713 \\
        Majority Vote & & 0.4306 & 0.4982 & 0.5267 \\
        Semantic reranking & Text & 0.1246 & 0.2562 & 0.3345 \\
        \cmidrule(lr){1-5}
        \multirow{2}{*}{RankGPT (Mistral)} & Text & $0.1495\pm0.0000$ & $0.2254\pm0.0017$ & $0.2894\pm0.0017$ \\
        & Text + G2T (Det. Lin.) & $0.1210\pm0.0000$ & $0.1815\pm0.0000$ & $0.2527\pm0.0000$ \\
        \cmidrule(lr){1-5}
        \multirow{2}{*}{RankGPT (Qwen2.5 7B)} & Text & $0.2562\pm0.0050$ & $0.3879\pm0.0058$ & $0.4591\pm0.0105$ \\
        & Text + G2T (Det. Lin.) & $0.2764\pm0.0044$ & $0.4128\pm0.0050$ & $0.4567\pm0.0034$ \\
        \cmidrule(lr){1-5}
        \multirow{5}{*}{Linear Regression} & Text & 0.2278 & 0.3345 & 0.3950 \\
        & Graph & 0.3238 & 0.4484 & 0.4911 \\
        & Text + Graph & 0.4021 & 0.4840 & 0.5338 \\
        & G2T (Det. Lin.) & 0.3167 & 0.4128 & 0.4733 \\
        & Text + G2T (Det. Lin.) & 0.3772 & 0.4733 & 0.5338 \\
        \midrule
        \multirow{5}{*}{Logistic Regression} & Text & 0.2242 & 0.3452 & 0.4164 \\
        & Graph & 0.3680 & 0.4990 & 0.5660 \\
        & Text + Graph & 0.3808 & 0.4769 & 0.5302 \\
        & G2T (Det. Lin.) & 0.3025 & 0.4270 & 0.4911 \\
        & Text + G2T (Det. Lin.) & 0.3915 & 0.4840 & 0.5302 \\
        \midrule
        \multirow{5}{*}{CatBoost} & Text & $0.1293\pm0.0017$ & $0.2230\pm0.0084$ & $0.3060\pm0.0029$ \\
        & Graph & $0.3665\pm0.0000$ & $0.4520\pm0.0029$ & $0.5089\pm0.0029$ \\
        & Text + Graph & $0.3820\pm0.0044$ & $0.4840\pm0.0058$ & $0.5409\pm0.0058$ \\
        & G2T (Det. Lin.) & $0.2112\pm0.0067$ & $0.2871\pm0.0102$ & $0.3713\pm0.0017$ \\
        & Text + G2T (Det. Lin.) & $0.3926\pm0.0034$ & $0.4840\pm0.0050$ & $0.5279\pm0.0034$ \\
        \midrule
        \multirow{2}{*}{MPNet} & Text & $0.2539\pm0.0044$ & $0.3381\pm0.0087$ & $0.4176\pm0.0044$ \\
        & Text + G2T (Det. Lin.) & $0.3642\pm0.0131$ & $0.4686\pm0.0131$ & $0.5302\pm0.0203$ \\
        \bottomrule
    \end{tabular}
\end{table}

\FloatBarrier

\section{Hardware setup for experiments}
\label{appx:hardware}
 Reproducing the described experiments can be difficult without enough hardware resources. The main steps with high resource consumption are subgraph preparation and LLM fine-tuning. To extract subgraphs for the dataset, we used the igraph library, which loads the entire knowledge graph into RAM for fast processing. For additional processing optimization, we applied multiprocessing techniques using the Python multiprocessing package, which requires enough CPU cores and  RAM for each process. According to our measurements, each process of subgraphs extraction requires at least 80GB of RAM and 1 CPU core. 
Another source of resource consumption is the fine-tuning of LLMs. The largest model in our experiments was the Mixtral model, which requires two Nvidia A100 GPUs with a total of 160GB of memory. 
Requirements for the hardware setup to reproduce experiments are presented in Table~\ref{tab:tech_requirements}.

\begin{table}[htbp]
    \setlength{\tabcolsep}{3pt}
    \caption{Hardware requirements for evaluation of presented experiments}
    \label{tab:tech_requirements}
    \fontsize{7pt}{9pt}\selectfont
    \begin{tabular}{l p{4cm} c}
        \toprule
        \textbf{Component} & \textbf{Requirements} & \textbf{Main consumer} \\
        \midrule
        CPU & AMD EPYC 7702 64-Core & Multithreading in graphs preprocessing \\
        RAM & 80GB (used by one thread, for multithreading 1TB was used) & Subgraphs extraction algorithm for loading knowledge graph into memory \\
        GPU & 2 x NVIDIA A100 & Training and evaluation of Mistral/Mixtral models \\
        \bottomrule
    \end{tabular}
\end{table}

\section{Entity linking for MKQA dataset}
\label{appx:mkqa_linking}

MKQA dataset contains entities only for answers to the questions unlike Mintaka, where entities presented in question can also be found in the original data. Entities from questions are required for of the first steps of our pipeline --- subgraphs extraction. To face with this issue we used custom entity linking solution. It works in the next way: firstly each question is provided to the LLM (openai/gpt-5-mini was used) with the next prompt: \textit{You are an expert in entity linking. For each provided sentence you must generate pairs in the format [('mention in text'\#'title of the entity');('mention in text of the another entity'\#'title of the another entity')] without any extra text. Example:\textbackslash nQuestion: Extract entities from the sentence: Hamilton wins race in Monaco?\textbackslash nAnswer: [(Hamilton\#Lewis Hamilton);(Monaco\#Monaco)]'. Extract entities from the sentence:}. 
The output of the LLM was parsed using regexp 
\begin{verbatim}
r'\[\([^)]+#[^)]+\)(?:;\s*\([^)]+#[^)]+\))*\]'
\end{verbatim} and split by ';' and '\#' symbols to extract titles of the detected entities. To find their Wikidata ids we used Wikidata text search via their official api and choose the first found entity for each request.

\end{appendices}

\clearpage

\bibliographystyle{ios1}           
\bibliography{bibliography}        

%

\end{document}